\documentclass[10pt,journal,compsoc]{IEEEtran}
\usepackage[margin=25mm]{geometry}
\usepackage{amsmath}
\usepackage{amsfonts}
\usepackage{amssymb}
\usepackage{graphicx}
\usepackage[utf8]{inputenc}
\usepackage{changepage}
\usepackage{tabularx}

\usepackage{verbatim}

\ifCLASSOPTIONcompsoc
  \usepackage[nocompress]{cite}
\else
  \usepackage{cite}
\fi
\ifCLASSINFOpdf
\else
\fi
\begin{document}
%
\title{A Critical Study on the Recent Deep Learning Based Semi-Supervised Video Anomaly Detection Methods}
%
%
%
%

\author{Mohammad~Baradaran,~\IEEEmembership{Graduate~Student~Member,~IEEE}
        
        ~Robert~Bergevin~\IEEEmembership{}


\thanks{The authors are with the Department of Electrical and Computer Science, Laval University, Quebec, Canada. E-mails: {mohammad.baradaran.1}@ulaval.ca, {robert.bergevin}@gel.ulaval.ca.}
}

%

%

\markboth{ }%
{}
%



\IEEEtitleabstractindextext{%
\begin{abstract}
Video anomaly detection is one of the hot research topics in computer vision nowadays, as abnormal events contain a high amount of information. Anomalies are one of the main detection targets in surveillance systems, usually needing real-time actions. Regarding the availability of labeled data for training (i.e., there is not enough labeled data for abnormalities), semi-supervised anomaly detection approaches have gained interest recently. This paper introduces the researchers of the field to a new perspective and reviews the recent deep-learning based semi-supervised video anomaly detection approaches, based on a common strategy they use for anomaly detection. Our goal is to help researchers develop more effective video anomaly detection methods. As the selection of a right Deep Neural Network plays an important role for several parts of this task, a quick comparative review on DNNs is prepared first. Unlike previous surveys, DNNs are reviewed from a spatiotemporal feature extraction viewpoint, customized for video anomaly detection. This part of the review can help researchers in this field select suitable networks for different parts of their methods. Moreover, some of the state-of-the-art anomaly detection methods, based on their detection strategy, are critically surveyed. The review provides a novel and deep look at existing methods and results in stating the shortcomings of these approaches, which can be a hint for future works.
\end{abstract}

\begin{IEEEkeywords}
Video anomaly detection, Spatio-Temporal feature extraction, Deep learning, Video analysis, Semi-supervised, Reconstruction and prediction.
\end{IEEEkeywords}}

\maketitle

\IEEEdisplaynontitleabstractindextext

%
\IEEEpeerreviewmaketitle

\IEEEraisesectionheading{\section{Introduction}\label{sec:introduction}}

%
%
%
%
\IEEEPARstart{A}{nomaly} Detection (AD) is one of the essential and crucial tasks in various applications, such as video surveillance, quality control in production lines, security systems in data transmissions, etc. Anomaly detection (a.k.a., abnormal event detection or outlier detection) involves detecting patterns in data (image, video, etc.) that do not conform to expected behavior or the notion of normal behavior (behavior conformed by the majority of data samples) \cite{b1}. In video anomaly detection, the goal is to precisely locate the anomalies (spatially and temporally) inside frame sequences. Anomalies may be of different types, but they generally share these assumptions: 1- Anomalies rarely take place (compared to normal events), so they have a low probability of occurrence. 2- Patterns of anomalies are distinct from normals (majority of the events). These assumptions are the keys to identifying anomalies, however, detecting anomalies is generally challenging for a number of reasons:\\
\\
1- There is not a limited and precise definition for abnormality. Anomaly patterns are diverse and unrestricted and hence cannot be modeled or predicted precisely.\\
2- The boundary between normals and anomalies is not often precisely defined. Besides, it is hard to classify the data instances near this boundary.\\
3- Abnormalities are highly contextual and their definitions can change considering the time, place and environment. For example, driving a car at a speed of 100 km/h is a normal behavior on a highway but it cannot be considered as normal, in a residential area.\\
4- Anomalies are rare (but diverse) and there are not enough labeled anomaly samples to train a model.\\
5- It is very difficult to define a precise boundary (model) around normals, which can cover all normal patterns and behaviors.\\
6- The most complex challenge would be intelligent anomalies (adversarial samples) which attempt to resemble normal patterns.

 

\subsection{Video Anomaly Detection (VAD)}


Video anomaly detection has the same definition as mentioned above, but here we deal with videos and we strive to detect anomalous video events, spatially and temporally. Hence, in video, appearance and motion are the key elements for defining anomalies and they should be extracted effectively and analyzed jointly. For video anomaly detection, we have the same general anomaly detection challenges (as mentioned above). There are some additional challenges, related to video analyzing, such as high dimensionality of video data, complex scenes, occlusions, high interaction inside video contents, low resolution, etc.

\subsection{Different video anomalies from data analysis viewpoint}
Anomalous data can be considered as one of these three subcategories, based on how they are analyzed: point anomalies, contextual (conditional) anomalies or collective (group) anomalies \cite{b1}. Depending on the condition and the context, video anomalies can belong to each of these categories, although mostly they should be considered as conditional anomalies, for a more robust and generalized performance.\\

For point anomalies, the anomaly is defined and recognized, by analyzing the value of one single data instance, individually. This value can be a scalar or a feature vector. For example, in a video, a frame can be labeled as an anomaly, simply by detecting an unexpected object (appearance feature vector) or by capturing a vehicle, which is moving with a speed greater than the allowed speed (regardless of the vehicle type or the place). For conditional anomalies, the contextual information is required, in addition to the value. In this case, a single factor (a variable value for instance) is not enough to make a robust and careful decision. For example, it is expected to see cars in the street, but in their designated lines, not on the sidewalk. As another example, although 100 km/h is an allowed speed on highways, it is considered to be an abnormal behavior on snowy slippy roads. As noted, in these examples, the values of the features are not individually enough and they can be interpreted differently, in different conditions (depending on the context). In collective anomalies, groups of data instances form the anomalies.  For example, the presence of one or a few people may be normal in the bank, but a group of one hundred people would be considered as an anomaly. Video anomaly can belong to any of the mentioned categories. However, conditional anomaly, generally, has a much more comprehensive definition, compared to point and collective anomalies and it can cover them both. Hence, it is more recommended to consider and interpret video anomalies as a conditional anomaly.

\subsection{Important points for an effective video anomaly detection}
Regarding the definition of video anomaly, different types of anomalies and challenges in video anomaly detection (VAD), the following items should be considered, to have an effective computer vision system for VAD.\\
\\
1: Precisely defining normality (normal patterns).\\
2: Extracting effective and discriminative spatiotemporal features, customized for the given task.\\
3: Considering the differences and similarities in and between normal and abnormal behaviors.\\
4: Considering environment information and its variations.\\

It is worth mentioning that anomaly detection approaches are divided into supervised, semi-supervised, unsupervised and weakly supervised methods. Semi-supervised methods have gained more attention, because the definition of anomalies is based on the definition of the normalities. In semi-supervised methods, it is assumed that there is enough labeled data to define normalities. On the other hand, there is not enough labeled data for anomalies (mostly because they are rare and difficult to capture and record and also they are diverse and it is hard to cover all anomalies). Important factors concerning semi-supervised anomaly detection that should be considered are:\\
\\
1) Availability of enough labeled data for normalities to cover all of the normal patterns.\\
2) Extracting compact and discriminative features for normal patterns, to ensure that normal features are very similar and close to each other and very distinct from features of  anomalies.
\\

The formulation of the problem, in this category, is as a one class classification (which may have multiple normal subclasses) and to learn a model, to express normal patterns (appearance and motion in video).

\subsection{Benchmark datasets popularly used for semi-supervised video anomaly detection}
UCSD (ped1 and ped2) \cite{b2} is one of the most popular datasets in semi-supervised video anomaly detection, in which the normal scenes include people walking in the walkways, while anomalies are due to the presence of unexpected objects in the scene (such as carts, bicycles, skate boards, etc.), different motion patterns (skateboard riding, etc.) and walking in the grass. The main challenge of this dataset is the low resolution of the frames. Moreover, in a few frames, a large number of people is observed in the scene. These scenes are rare and compatible with the definition of anomaly; however, these frames are considered as normal and not annotated as an anomaly in the dataset. Hence, it is challenging for the system to perceive the scenes involving a large number of people as a regular scene (since this is not a frequently occurring scene). The Shanghai Tech dataset \cite{b3} is a similar dataset to UCSD, from the viewpoint of definition of anomalies and normals. However, unlike The UCSD dataset, its resolution is high and it would be easier for systems to recognize anomalous objects, using appearance based features.\\

The CUHK Avenue dataset \cite{b4}, UMN dataset \cite{b5} and Subway dataset \cite{b6} consider the activities of people in quite different scenarios. The Street Scene dataset \cite{b7} is a recently proposed high resolution dataset and, among all mentioned datasets, it is the most challenging, since: 1) the events, inside, are highly contextual (in addition to the appearance and motion information, the location information of the objects are essential in defining the anomalies). Figure 1 illustrates this point. 2) The anomalies are of various types and they are numerous. The mentioned datasets are compared in detail, in Table 1 and Table 2.

\begin{figure}[htp]
    \centering
    \includegraphics[width=8cm]{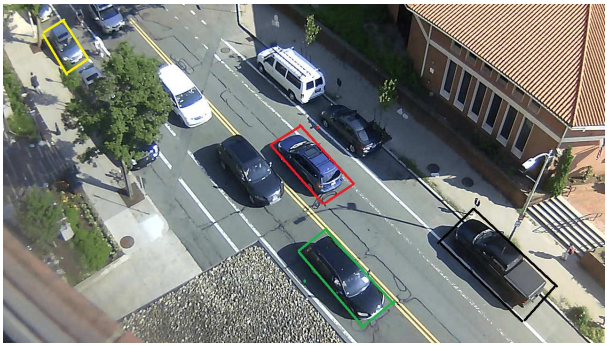}
    \caption{A frame from the Street Scene dataset \cite{b7}. Anomalies in this dataset are highly contextual. For example, the definition of anomaly is different, in these 4 positions (4 different boxes), for the same class of object. Best viewed in color.}
    \label{fig:street}
\end{figure}

\begin{table*}[t]

 \caption{Detailed information on benchmark video anomaly detection datasets.}

{\footnotesize
 \begin{adjustwidth}{-1cm}{}
\begin{tabular}{|p{2.5cm} ||p{1.6cm} |p{1.6cm} |p{1.6cm} |p{1.6cm}| p{1.6cm}| p{1.6cm}| p{1.3cm} |p{1.3cm}|}
\hline
Datasets& Training frames & Test frames& Resolution& NO. of anomalies& \footnotesize {Annotation} & Color& No. of scenes& Format\\
\hline
\hline
UCSD-ped1 & 6,800 & 7,200 & 238 x 158 & 54 (5 types) & Pixel based-binary mask & gray & 1 & tif \\
\hline

UCSD-ped2 & 2,550 & 2,010 & 360 x 240 & 23 (5 types) & Pixel based- binary mask & gray &1 &tif\\
\hline
ShanghaiTech & 274,515 & 42,883 & 856 x 480 & 130 & Pixel based- binary mask & color& 13 & avi \\
\hline
Street scene & 56,847 & 146,410 & 1280 x 720 & 205 (17 types) & Pixel based- bounding box & color & 1 & jpg \\
\hline
CUHK Avenue & 15,328 & 15,324 & 640 x 360 & 47 (5 types) & Pixel based-mask & color & 1 & tif \\
\hline
UMN & 7,740 total frames &  - & 240 x 320 & 11 (1 type) & Frame based & color/ gray & 3 & avi \\
\hline
Subway (Entrance) & 18,000 & 68,535 & 512 x 384 & 66 (5 types) & Frame based & gray & 1 & avi,tif \\
\hline
Subway (Exit) & 4,500 & 34,440 & 512 x 384 & 19 (3 types) & Frame based & gray & 1 & avi,tif
\\
\hline

\end{tabular}
 \end{adjustwidth}
 }

\end{table*}

\begin{table*}[t]
 \caption{Comparing existing benchmark VAD datasets, based on their definition of anomaly and their challenges. Different items are separated by asterisks (*) in the table.}

{\footnotesize
 \begin{adjustwidth}{0.7cm}{}
\begin{tabular}{|p{2.5cm} ||p{5.6cm}| p{5.6cm}| }

\hline
\textbf{Datasets} & \textbf{Anomalies (some)} & \textbf{challenges/ special points} \\
\hline
\hline

\textbf{UCSD} & Non pedestrian entities in walkways (bikes, skates, small carts, wheelchair),
anomalous pedestrian motion patterns (people walking across the walkway or on the grass). & *Definition of anomaly for Ped2 is the same with Ped1
*Different scales for different distances.
*Object types are not always recognizable (due to resolution and distance).
 \\
\hline

\textbf{ShanghaiTech Campus} & Sudden motion, such as chasing and brawling, unexpected objects. & *Multiple scenes with multiple view angles.
*Complex lighting conditions.\\

\hline

\textbf{Street scene} & Jaywalking, loitering, car outside lane, car u-turn, car illegally parked, biker on sidewalk, etc. (17 types of anomalies) & *The anomalies are highly contextual and more challenging than other datasets
*High resolution
*High number of anomaly types.
*Presence of minor camera motion in some frames.
\\
\hline

\textbf{CUHK Avenue} & Throwing objects, loitering, running.&
*The size of people may change because of the camera position
and angle
*Camera shakes in some frames.
\\
\hline

\textbf{UMN} & Crowd escaping quickly from the scene.&
*Number of anomalies is limited (just one anomaly type)
*The video is short.
*Low resolution.
\\
\hline

\textbf{Subway} & Moving in a wrong direction, entering without payment, loitering.&
*Noisy video
*There is a big timer on the screen. *Objects at distance are not clear.
\\
\hline

\end{tabular}

 \end{adjustwidth}
 }

\end{table*}

\subsection{Other surveys}
There are a few other survey articles published concerning anomaly detection. For example, Ramachandra \cite{b8} surveys single-view video anomaly detection methods, with a special consideration of the applicability of methods on currently available benchmark datasets. Kiran \cite{b9} reviews the state-of-the-art Deep Learning (DL) based approaches for anomaly detection in videos and categorizes them based on the criteria of detection and the type of network used. As it reviews reconstruction-based, prediction-based and generative models for anomaly detection, it is similar to our work, in some parts. However, the deep networks are studied mostly focusing on their structures and basic concepts, but not from a feature extraction viewpoint and their compatibility with the anomaly detection task. Raghavendra \cite{b10} presents a general review on DL-based anomaly detection methods and reviews their applicability in different fields of application (not limited to video, but also other applications such as fraud detection, medical anomaly detection, sensor networks, etc.). This review is an application-based categorization and mostly summarizes the Deep Neural Network (DNN) types, used for various applications. Chandola \cite{b1} reviews Machine Learning based anomaly detection methods (conventional Machine Learning (ML) approaches) based on the different pattern recognition techniques used (such as clustering, classification, neural networks, etc.) and it studies them for different applications. Moreover, this review explains the basic assumption, advantages, computational cost, etc, for each of the techniques. Bulusu \cite{b11} has provided a review on DL-based anomalous instance detection methods. Its focus is on discussing unintentional and intentional anomalies, specifically in the context of DNNs. Shibin \cite{b12} reviews evaluation metrics and popular evaluation schemes, used to measure the performance of video and image anomaly detection approaches. These surveys, as well as a few others, have completely compared anomaly detection methods based on the amount of supervision (supervised, unsupervised and semi-supervised methods), discussing their advantages and disadvantages, hence this will not be discussed in this paper. Table 3 compares existing surveys, based on the subjects they have covered.

\begin{table*}[!hbt]
 \caption{Comparing existing AD and VAD surveys, based on their contents and subjects covered. DNN stands for Deep Neural Network, DL for Deep Learning, ST for Spatio-Temporal and ML for Machine Learning. Different items are separated by asterisks (*) in the table.}

{\footnotesize
 \begin{adjustwidth}{-1.40cm}{}
\begin{tabular}{|p{3.5cm} ||p{1.7cm}| p{1.7cm}|p{1.7cm}| p{1.7cm}| p{1.7cm}| p{1.7cm}| p{1.9cm}|   }

\hline
\textbf{Reference} & \textbf{[1]} & \textbf{[8]} & \textbf{[9]}& \textbf{[10]}& \textbf{[11]}& \textbf{[12]}& \textbf{Our survey}\\
\hline
\hline

\textbf{Reconstruction} & & \checkmark & \checkmark & & & & \checkmark\\
\hline

\textbf{Prediction} & &  & \checkmark & & & & \checkmark\\
\hline

\textbf{Object centric} & &  &  & & & & \checkmark\\
\hline

\textbf{Segmentation} & &  &  & & & & \checkmark\\
\hline

\textbf{Memorization} & &  &  & & & & \checkmark\\
\hline 

\textbf{ST feature extraction} & &  & \checkmark & & & & \checkmark\\
\hline

\textbf{Datasets} & & \checkmark & & & & & \checkmark\\
\hline

\textbf{Evaluation metrics} & & \checkmark & \checkmark & & & \checkmark& \checkmark\\
\hline

\textbf{DNN/ ML approaches} & ML & DNN/ML & DNN & DNN & DNN/ML &  & DNNs \\
\hline

\textbf{Focused applications} & Intrusion detection,  fraud detection, medical anomaly detection, industrial damage detection, image processing, text, sensor networks. & Video anomalies & Video anomalies & Fraud detection, cyber intrusion detection, IOT, video, industrial damage, sensor, etc. & Fraud detection, malware detection, healthcare, video surveillance, etc. & Image and video anomaly detection & Video anomalies\\
\hline

\textbf{Topics covered} & *Anomaly detection in different applications. *Different ML approaches for anomaly detection. & *Distance-based, probabilistic and reconstruction-based anomaly detection approaches. & *Reconstruction models, predictive models, generative models for AD. & Semi-supervised, unsupervised, hybrid models, one class neural networks for AD. &*Detection of unintentional anomalies, *Detection of intentional anomalies*Applications & *Evaluation schemes. *Evaluation metrics. & *DNNs from a feature extraction viewpoint. *AD methods based on their ST feature extraction processes. *Semi-supervised AD methods.
\\
\hline

\end{tabular}

 \end{adjustwidth}
 }

\end{table*}

\subsection{Contributions}
In this survey:\\
\\
- Deep Neural Networks are reviewed and compared, from the point of view of spatiotemporal feature extraction and pattern learning. This novel viewpoint can be helpful in video anomaly detection.\\
- Recent DL-based semi-supervised anomaly detection approaches are reviewed and compared, stating their strong and weak points.\\
- Common aspects of all recent DL-based semi-supervised anomaly detection approaches (especially, their implicitly common strategy for anomaly detection) are stated. This provides a new, global and integrated perspective to the field.\\
- Most recent proposed DL-based semi-supervised anomaly detection approaches are covered.\\
- Selected experiments are conducted to illustrate the strong and weak points of some of the video anomaly detection methods.

\subsection{Organization of the paper}
The rest of this article is organized as follows (see Figure 2). In Section 2, deep neural networks are reviewed from different points of view, analyzing their applicability for several steps of video anomaly detection (this section also contains some examples used in supervised or unsupervised anomaly detection methods, to clarify the subject). In section 3, a general look at different anomaly detection approaches is provided and some of the recent state-of-the-art DL based semi-supervised anomaly detection methods are presented and compared, based on how they formulate and address the problem and finally, shortcomings of existing methods are listed, which can be the subject of future work.

\begin{figure*}[htp]
    \centering
    \includegraphics[width=12cm]{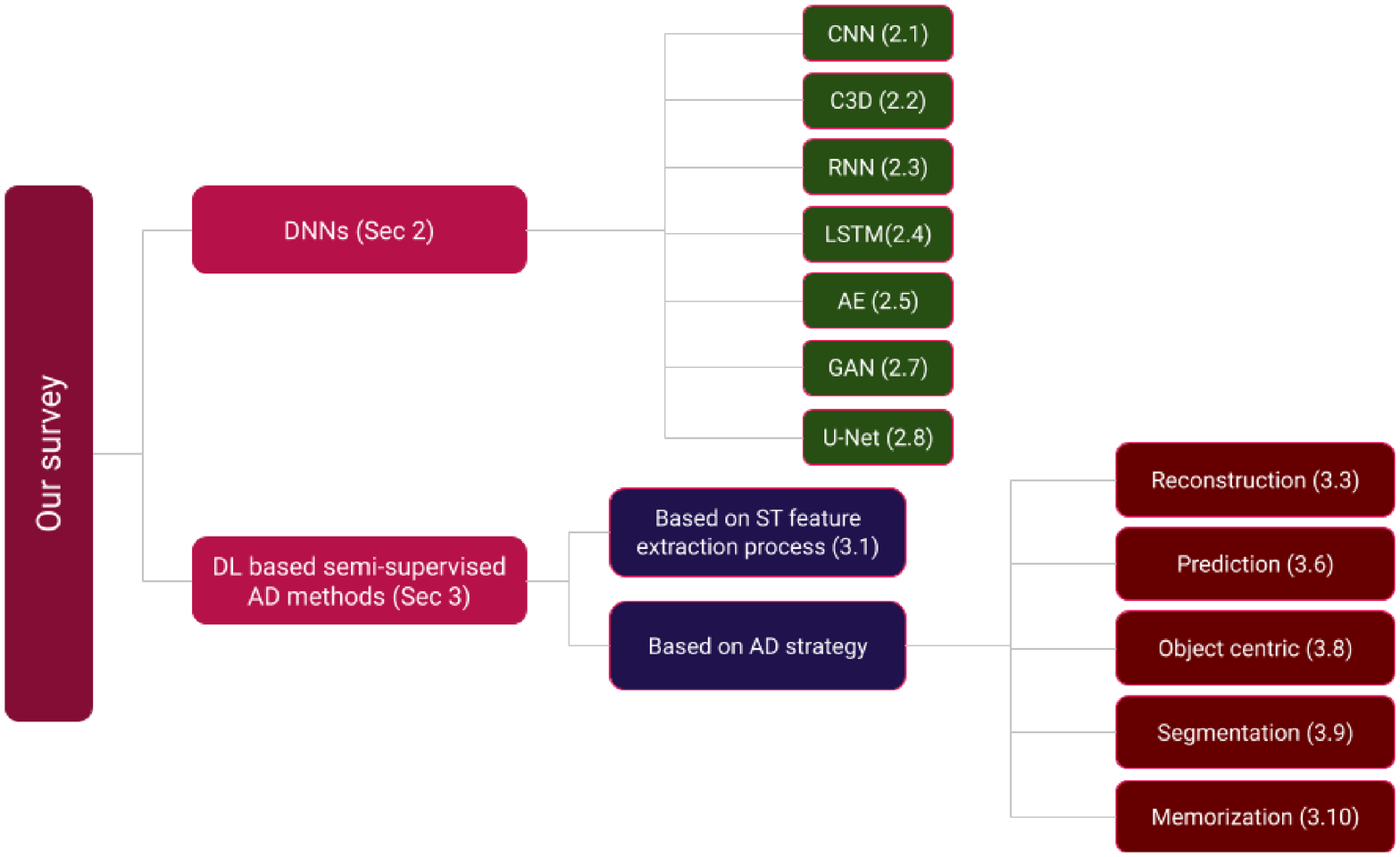}
    \caption{Illustration of different sections in our survey (DL: Deep Learning, AD: Anomaly Detection, ST: Spatio-Temporal). Numbers in each block show the related section in the paper.}
    \label{fig:chart}
\end{figure*}

\section{Deep Neural Networks}
Like most computer vision tasks, video anomaly detection is completely reliant on effective feature extraction. Hence, it is very important to have a good understanding of DNNs feature extraction capabilities, as they are the key tool for feature extraction and pattern learning (in DL-based approaches). In this section, a general, yet effective, look at various deep models and their applicability for different related sub-tasks is provided, and their compatibility with different data types is analyzed.\\

Deep learning has brought great success to various applications and research fields, especially to computer vision applications, in analyzing high-dimensional data. DNNs have been useful for different purposes and steps in computer vision applications. More specifically, in video anomaly detection, they have been used to:\\
\\
1) Extract discriminative high-level Spatio-Temporal features, for different types of data (such as spatial data, sequences, etc.), by using proper architectures (such as CNNs, RNNs, etc.)\\
2) Learn, model and memorize patterns and information.\\
3) Formulate solutions to problems associated with different tasks.\\
4) Differentiate between normal and abnormal patterns.\\

Hence, it will be useful to briefly review DNNs, considering the mentioned factors. These networks are analyzed from different points of view such as their architectures, feature extraction ability, compatibility with different data types and their applicability to different tasks.

\subsection{Convolutional Neural Networks (CNN)}
Convolutional neural networks are special forms of feed-forward neural networks and are composed of multiple convolutional and pooling layers, which are followed by a few Fully Connected (FC) layers, at the end of the network. Unlike the Fully Connected Networks, the architecture of CNNs is compatible with 2D structured inputs (such as images or any other 2D signal), which helps effectively preserve the spatial structure of inputs. Feichtenhofer et al. \cite{b13}, present a deep insight into convolutional neural networks, for video recognition tasks. Convolutional layers are composed of multiple kernels, which are convolved with the input image or mid-layer activation maps to produce next-level activations. Convolutional layers benefit from several advantages, as indicated below:\\
\\
- The weight-sharing mechanism, in each mapping process, has intensively reduced the numbers of parameters.\\
- Convolution makes the network robust and invariant to translation.\\
- A convolutional layer attempts to identify local patterns in the input.\\

In some networks such as Autoencoders, in order to reconstruct an image from extracted features in the latent space, there must be up-sampling-like layers to increase the resolution of feature maps. Transpose convolution, which is also referred to as deconvolution or up-convolution, is a convolution-based operation, which increases the resolution of its input. Up-sampling is a similar operation to transpose-convolution, but the main difference is that transpose convolution has trainable kernels.

\subsubsection{Characteristics of CNNs}
Videos are consecutive frames that should be processed both separately (image processing) and also in connection to each other (considering their temporal dependencies). Convolutional neural networks are generally the essential elements for image processing. This is due to some important characteristics of CNNs, which make image processing more effective, efficient and even less challenging. Some of these characteristics are listed as below:\\
\\
1: Reduced number of parameters: Thanks to local connectivity and shared weights, CNNs have much less parameters, compared to Fully Connected Networks, and hence are easier to train.\\
2: Shift/Translation invariance: this means that by any shift in input, the result does not change (because of convolutional and pooling layers).\\
3: Transfer Learning: Transfer Learning possibility is one of the strengths of CNNs, in which pre-trained networks are used for feature extraction, in other similar datasets. Nazare et al. \cite{b14} studies the quality of features, extracted by pre-trained CNNs, for anomaly detection tasks. Ionescu et al. \cite{b14.1} and Aburakhia et al. \cite{b14.2} is a good example for the application of pre-trained CNNs for extracting appearance features to detect anomalies in videos.\\ 
4: Convolutional neural networks (CNNs) are suitable for processing an input data that has an inherent grid-like topology.\\
5: CNNs extract rich features at different semantic levels.\\
6: Convolutional neural networks (ConvNets) are biologically inspired.\\
7: By removing the FC layer from CNN, the restriction on image size can be removed.\\
8: More filters capture more features but increase the computational cost \cite{b15}.

\subsubsection{CNNs from a feature extraction point of view}
As mentioned in the previous subsection, CNNs render image processing (and hence video processing) more efficient and effective, due to their ability to exploit spatial features. Therefore they are the prime element for spatial feature extraction from frames. Here are some important aspects to consider regarding CNNs from the feature extraction viewpoint:\\
\\
1: Experiments show that features in the first layers are low-level and local. For example, filters in the first layer are edge detectors and color filters. The edge detectors are at different angles and allow the network to construct more complex features in the next layers \cite{b14}.\\
2: Layers towards the end of the network learn high-level combinations of the features learned in the earlier layers (see Figure 3).\\
3: Although the deeper layers have higher-level features and are usually used as feature representation,  in order to have a better performance for a special task, it is the target task which precisely defines the layer from which the features should be extracted. For example, in some tasks, such as iris recognition, the recognition accuracy drops after special layers, because the network captures only the abstract and high-level information and it does not distinguish much between diverse iris patterns \cite{b16}.\\
4: The feature extraction process in a CNN is adapted according to the class of the input image.\\
5: Reducing the kernel size can improve the capture of smaller details in the picture, while missing the global information in the frame and may result in greater confusion. Larger kernels, on the other hand, will lead to a global look at the image, while missing the details \cite{b17}. This is extremely important when there are objects at different distances from the camera (different scales). Hence, the filter size should be selected considering the task, dataset and the application. Some works, such as \cite{b27}, utilize inception modules in early layers, to automatically select the proper kernel size.\\
6: Each kernel is in charge of learning special features from the image. For example, convolutional kernels are capable of capturing features such as edge, line, texture, shape, intensity, color, etc. \cite{b18}.\\
7: Deeper networks extract more complex features.\\
8: Earlier layers in a CNN concentrate on generic features (independent of the task), while deeper layers extract features more specific to the problem and the goal.

\begin{figure*}[htp]
    \centering
    \includegraphics[width=12cm]{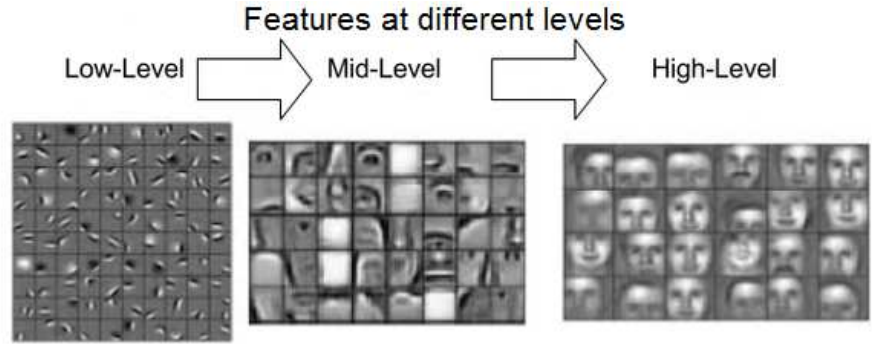}
    \caption{Different levels of features extracted in CNNs. In this figure, different levels of features are extracted for human face pictures. Left: extracted low level features are generic and focusing on edges. Middle: CNN focuses on different parts of the object at mid levels. Right: deeper layers provide a global look at objects, extracting high level features. This figure originally appeared in \cite{b18.1}.}
    \label{fig:cnn}
\end{figure*}

\subsubsection{CNNs for spatiotemporal feature extraction}
As explained before, CNNs are excellent and powerful in feature extraction from images. When it comes to consecutive frames (video clips or other 3D tensors), CNNs are not, by nature, suitable for the capture of temporal patterns \cite{b19}, since they consider single frames as input. Moreover, 2D convolutional kernels map each receptive field (2D or 3D) to one channel (note that, kernel depth in CNNs is equal to the number of input channels). In order to allow the network to be aware of temporal variations, the use of a cuboid of frames (instead of a single frame), as the input, has been proposed. However, the first convolution destroys the temporal structure and does not show promising results in capturing temporal patterns \cite{b19.1}.

\subsection{3D Convolutional Networks (C3D)}
In C3D, 3D kernels (with sizes smaller than the width, height and depth of frame sequences) are applied on consecutive frames (usually 16 frame clips) and the output is a 3D tensor, unlike the 2D networks which produce a one channel output for either an image input or sequence of frames (see Figure 4). The C3D network considers spatial information in the first few frames and it starts to consider the temporal information in the following frames \cite{b20}.

\begin{figure}[htp]
    \centering
    \includegraphics[width=6cm]{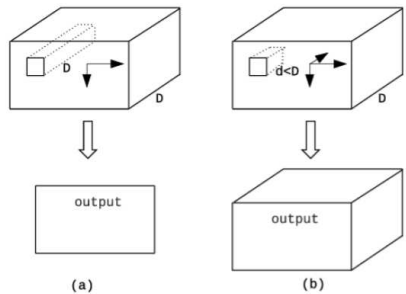}
    \caption{Different structures in  2D and 3D convolutions and their different output feature maps \cite{b21}. (a): in a 2D CNN, the output feature map is a single channel tensor. (b): in a 3D CNN the output is a 3D tensor.}
    \label{fig:c3d}
\end{figure}

\subsubsection{Characteristics of C3Ds}
The architecture of C3Ds seems to be a good choice for spatio-temporal feature extraction. However, there are some points about C3Ds that should be considered for feature extraction.\\
\\
1- C3D achieves better results in video analyzing tasks (such as video classification and video retrieval), compared to 2D CNN, as it captures both spatial and temporal information \cite{b22}.\\
2: C3D requires a high number of parameters, thus it is computationally expensive and difficult to train, which makes it prone to overfitting \cite{b23}.\\
3- Modeling the long sequences is not addressed in C3D, because it leads to a huge computational cost.\\
4- C3D cannot take advantage of Transfer Learning effectively (unlike 2D CNNs). Although some methods have been proposed to cover this issue partially, the results are not as good as those obtained in 2D CNNs.

\subsection{Recurrent Neural Networks (RNN)}
Basic feed-forward networks (such as CNNs) accept a fixed-sized input and produce fixed-sized outputs. This is one of the shortcomings of the feed-forward networks, which are therefore not applicable for some applications, such as language translating or frame captioning, in which the length of the input sentence (or image) and its translation might be different. This problem is addressed by recurrent neural networks. Moreover, unlike a feed-forward network, in which data pass through layers once, in RNNs, they cycle in a loop and touch neurons several times. In this way, RNNs not only consider the current input, but also care about its temporal neighbors (past or future frames). More importantly, as Recurrent Neural Networks feature inner loops, they allow the information to persist \cite{b24}.

\subsubsection{Characteristics of RNNs}
Recurrent neural networks are, by nature, compatible with sequences \cite{b24.1}. Hence they are widely used for temporal feature extraction. However, regarding the models' abilities and the target task, some points should be considered regarding RNNs, as listed below:\\
\\
1: As they benefit both new input data and the previous hidden state, as the input to the network, they are able to model sequences and to extract temporal information.\\
2: Thanks to the presence of the hidden state, they benefit from an internal memory \cite{b24.2}.\\
3: This model is not limited to fixed input and output sizes and hence is appropriate for several tasks such as video captioning, translating, etc.\\
4: RNNs have difficulties learning long-term dependencies, because of vanishing and exploding gradients \cite{b24.3}.

\subsection{Long Short-Term Memory Units (LSTMs)}
LSTM is a special type of RNN, designed to avoid the long-term dependency problem. LSTMs are gated memory blocks, which include 3 special gates in their chain-like structure, and in addition to hidden states (as is in RNNs), they have cell states. Carefully regulated by gates, LSTM has the ability to remove or add more information to the cell states. Gates, in LSTM, are composed of a sigmoid layer and a pointwise multiplication operation. Since the sigmoid function produces outputs between zero and one, it defines how much information should be deleted or passed \cite{b25}.\\

Like every type of neural network, layer size (memory units, here in LSTM) and network depth are the hyper-parameters to choose. Generally, deeper models show better performance in extracting richer features, compared to shallow models, but using much deeper models does not always guarantee a best performance, for all types of applications and tasks.

\subsubsection{Characteristics of LSTMs}
LSTM has some extra advantages compared to simple RNNs in modeling sequences (for example an inherent memory), which makes it the first choice for sequences, in most cases. However, other practical points, as listed below, should be considered about LSTMs.\\
\\
1: LSTMs handle exploding and vanishing gradients effectively, thus they are able to model longer sequences, compared to the basic RNN structure, although this length also depends on the nature of the sequence data and its inner correlation \cite{b15}.\\
2: Although LSTMs have no difficulty in modeling long dependencies, they lead to high computational complexity, when modeling long sequences.\\
3: LSTM is the basic element of temporal attention mechanisms.\\
4: LSTMs have the ability to learn the context required for making predictions in sequence data, therefore they are widely used for forecasting tasks \cite{b25.01}.

\subsubsection{C3D versus LSTMs, in modeling temporal information}
Although both 3D convolutional networks and recurrent neural networks consider sequences and model temporal information, the nature of patterns, captured by these models, are quite different. In LSTMs, based on the task, the network can be encouraged to select meaningful time dependencies and forget unnecessary items. Moreover, the network follows the evolution between sequences. In C3Ds, the network attempts to memorize the patterns inside the training cuboid (frame sequences) without explicitly emphasizing the order of the frames. Moreover, the extracted patterns in C3Ds are more generic \cite{b20}.

\subsubsection{Special points regarding ConvLSTM}

LSTMs in their basic form are not suitable for 2D spatially structured data. Hence they are extended to ConvLSTM, in which multiplications are replaced with convolutions.\\
\\
1: ConvLSTM shows great performance in extraction of spatiotemporal features, by taking advantage of its two main elements: i- LSTM to capture long temporal dependency and ii- Convolution for structured spatial information.\\
2: Due to convolution, ConvLSTM is capable of capturing local spatial information and suitable for spatiotemporal localization tasks \cite{b25.09}.\\
3: In LSTM, convolutional kernel size of input-to-input connection, defines the resolution of the feature map produced from the input. In addition, the filter size of hidden-to-hidden connections defines the collective information from previous steps. Moreover, larger transitional kernels capture faster motions, while smaller kernels perceive slower ones \cite{b19}.

\subsubsection{GRU versus LSTM}

The Gated Recurrent Network (GRU) is another improved version of the standard recurrent neural network to solve the vanishing gradient problem of a standard RNN. It is a gated memory block similar to LSTM, with a different number of gates (3 gates in LSTM and 2 in GRU). Chung et al. \cite{b25.1} evaluate the performance of these networks on sequence modeling.\\
\\
1: GRU has a simpler architecture compared to LSTM and its training is faster \cite{b25.05}.\\
2: In theory, LSTMs can learn longer sequences than GRUs and they perform better for longer sequences.\\
3: In general, the performance of GRU is comparable to that of LSTM. However, the relative performance of each method depends on the data and the application \cite{b25.1}.
4: Unlike GRU, which exposes its full content (seen or used content) without any control, the amount of the memory content is controlled by the output gate, in LSTM \cite{b25.1}.

\subsection{Autoencoders}
A deep Autoencoder is an unsupervised learning network architecture (learning from unlabeled training samples) composed of two main sections, encoder and decoder, which aims to map input data to a latent space, in order to extract deep features and then reconstruct the input using extracted features. In other words, it attempts to learn an approximation to the identity function, so that the output would be similar to the input. Recently, autoencoders are widely used in anomaly detection (especially video anomaly detection). This is because of their  ability in unsupervised representation learning. Here are some points regarding Autoencoders (AEs), which should be considered by researchers, in video anomaly detection:\\
\\
1: They extract effective representations from data, in an unsupervised approach.\\
2: Autoencoders are be effectively used for noise removal \cite{b25.2}.\\
3: Autoencoders are effectively used for dimensionality reduction similar to PCA. The difference is that PCA is restricted to a linear map, while auto encoders can have nonlinear encoders/decoders \cite{b25.3}.\\
4: A basic Autoenoder is not suitable for image processing, as it flattens the image to a vector and destroys the spatial structure (this problem is addressed by convolutional Autoencoders).\\
5: A Baseline (vanilla) Autoencoder is composed of fully connected layers and hence is computationally expensive.\\
6: A baseline Autoencoder is not complex enough to learn complex information (such as image content), and thus generally attempts to memorize and average the data (this problem is addressed, partially, by Variational Autoencoders). \\
7: The fundamental problem with Autoencoders, for generation tasks, is that their latent space (where they convert their inputs to), may not be continuous, or allow easy interpolation \cite{b26}.\\
8: It is challenging to select the best compression degree for Autoencoders.

\subsubsection{Shortcomings of deep Auto-encoders}
As mentioned before, AEs are appealing tools for anomaly detection researchers. However, they sometimes do not produce the desired results, mostly due to these facts: \\
\\
1: They are prone to vanishing gradients.\\
2: They reproduce a lower-quality version of the input image, without explicitly considering its high level contents \cite{b26.05}.\\
3: Autoencoders confront information imbalance in each layer \cite{b3}.\\
4: Autoencoders are unsupervised feature extractors and are not aware of the classes of the objects inside the image.\\
5- Autoencoders suffer from memorization and their reconstructed images are blurry \cite{b26.1}. 

\subsection{Variational Autoencoders (VAE)}

Variational AE (VAE) is a generative variant of classical AEs, which assumes a probability distribution (such as a Gaussian distribution) for the source input data and it attempts to capture the parameters of the distribution, through an encoding-decoding process. In VAE, not only does the network attempt to reconstruct an image, but the network is also asked to consider the same distribution, for generation of new samples, as it was in the training dataset. Important characteristics of VAEs can be listed as below:\\
\\
1- VAEs produce a lower-dimensional representation of the input data (like classical AEs).\\
2- By design, VAEs have continuous latent spaces, which makes random sampling and also interpolation easier \cite{b26.2}.

\subsection{Generative Adversarial Network (GAN)}
GANs are a set of generative networks, which are able to generate new contents. In Generative Adversarial Networks (GANs), the aim is to produce new data (such as images) which look real. In fact, this goal is a min-max game between a Generator (G) and a Discriminator (D), so that D tries to recognize real and unreal images, while G tries to produce images which look real. This learning architecture gives these networks a good ability, suitable for frame processing tasks, some of which are listed below:\\
\\
1- GANs allow CNN to learn an implicit distribution from data patterns \cite{b27}.\\
2- GAN has a good applicability for video prediction \cite{b3}.\\
3- GANs are used to produce data for prediction applications, in which not enough training data is available \cite{b28}.\\
4- GANs produce sharper images compared to VAEs \cite{b29}.

\subsubsection{Main challenges of GANs}
Despite GANs special abilities in feature extraction and frame processing, there are some challenges, specific to these networks, which sometimes lead to reduction in their use. The most noticeable challenges are:\\
\\
1- They require a precise selection of hyper-parameters.\\
2- They need multiple initializations \cite{b30}.\\
3- It is difficult to train adversarial methods, such as GANs.

\subsection{U-Net}
Autoencoders suffer from vanishing gradients and lack of information symmetry in their architectures. To tackle this problem, U-Net is proposed, which adds a shortcut between a high-level layer and a low-level layer with the same resolution \cite{b31}. The difference between U-Net and AE, in architecture, is illustrated in Figure 5.

\begin{figure*}[htp]
    \centering
    \includegraphics[width=12cm]{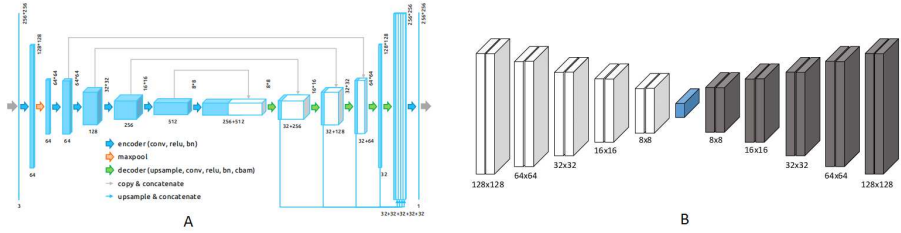}
    \caption{Illustration of similarity and difference, in architectures, between U-Net (A) and Conv-AE (B). Figures A and B originally appeared in \cite{b31.1} and \cite{b31.2}, respectively.}
    \label{fig:unet}
\end{figure*}

\section{Anomaly Detection methods}
In this section, anomaly detection methods are reviewed from two different viewpoints. First, the methods are reviewed, based on how they jointly extract spatiotemporal features. Then, different methods are studied, considering how they formulate the task and approach the problem.

\subsection{Methods based on Spatio-temporal (ST) feature extraction}

A video is a sequence of frames, which are evolved over time. Therefore, the two main important defining attributes are appearance and motion, from which video analysis is performed. Appearance is the first attribute which attracts the attention of the analyzers. Anomalies in videos can be due to the presence of unknown (previously unseen) objects, which can be defined by appearance-based features. However, this is not the only factor for the definition or creation of anomalies in videos. In a variety of cases, it is the motion, which defines the anomaly. For example, an irregular speed of a car inside a street can determine an anomaly taking place in that scene. Similar to appearance, motion features should be analyzed and modeled both locally and globally, in order to gain a better understanding of the contents of the video. Motion patterns can be represented to the network directly by motion-based features (such as optical flow) or they can be captured by sequence aware networks (such as the RNN family). The importance of considering motion is that most of the anomalies, in the real world, take place with moving objects. Humans consider and analyze both appearance and motion factors jointly and interactively, because, generally, motion and appearance are not always independent, but each one can also be a support to determine the other. For example, the motion pattern of an object (let us assume a snake here) can be a support for its recognition, in addition to its appearance features (such as shape, color, etc.). Various methods and models are proposed for spatiotemporal feature extraction (considering both motion and appearance simultaneously), which were studied in the previous section (literature on deep learning based models). However, from other viewpoint (the process of jointly extracting motion and appearance features), methods can be categorized into the following categories:\\
\\
A- Single model-Single path methods: in this category (e.g., \cite{b33,b34,b35}), spatiotemporal features are extracted through a single path (single branch) process, by a single model. The most noticeable type of model in this category is C3D, which is able to extract rich spatiotemporal features for different applications, especially action recognition. However, a few important points should be considered regarding this model, as listed below:\\
- There are no effective pre-trained C3D networks (like pre-trained CNNs).\\
- C3Ds are difficult to train (high computational cost) and require an enormous amount of training data \cite{b35.01}.\\
- C3Ds capture local motion patterns \cite{b35.02}.\\
\\
B- Two stream methods: In these methods (e.g., \cite{b35.1,b35.2}), motion and appearance are modeled separately, using two separate but usually identical branches. Generally, the input of one of the branches is a raw frame and this branch is in charge of modeling appearance, through a frame reconstruction or prediction task. In a complementary manner, the second branch attempts to capture and model motion patterns. This is generally achieved by receiving an explicitly-extracted motion feature (e.g., optical flow map) and modeling it through a reconstruction task \cite{b35.3}, or by getting a raw frame and learning the associated motion patterns by predicting its corresponding optical flow map (i.e., through an image translation task). The two branches of the model are usually optimized jointly, which implicitly encourages the model to learn the features of both types in an integrated way. To better integrate motion and appearance, some methods such as \cite{b36}, propose to add cross-branch connections, to transfer more information between the branches. Nguyen et al. \cite{b27} use a similar strategy (but in a novel way), to jointly extract Spatio-Temporal features. In their approach, two identical but separate branches (i.e, decoders, in this example) decode the extracted features of a frame (produced by a common encoder), consecutively to reconstruct the input frame, and to estimate the optical flow. In the inference stage, they compute the reconstruction/prediction error of each branch in order to detect anomalies.\\
\\
C- Hybrid methods: in methods of this category (such as \cite{b15}), generally multiple networks (each specialized in extracting specific features, such as motion and appearance) are connected in order to extract spatiotemporal features. As numerous research studies have proved, CNNs are powerful in image analysis and, on the other hand, RNN families are, by nature, suitable for analyzing video sequences. Hence several methods with different architectures have connected these two types of networks to extract suitable spatiotemporal features for anomaly detection.\\

It is worth mentioning that several fusion approaches, in several levels such as pixel-level (concatenation before ST feature extraction), feature-level (fusion of features before decision making), and score-level (fusion of anomaly scores extracted from different features) have been proposed to combine effects of different features, for a better anomaly detection \cite{b35.3}.

\subsection{Common approach of DL-based semi-supervised VAD methods}
Various DL-based semi-supervised VAD methods have been proposed, which use different strategies for anomaly detection. They have approached the problem by performing various tasks (reconstruction, segmentation, prediction, etc.), which are not of direct interest \cite{b36.01} and seem to be apparently unrelated to the task of anomaly detection. All of these different methods, as a self supervised task, mine different features \cite{b36.02}, however they all exploit the fact that all machine learning (ML) methods generally achieve the desired results for the data types on which they are trained (or at least for similar enough data types). In other words, ML methods do not guarantee that the desired results will be produced for a test data, which is new and different from the training set. In the following sections, the methods are all based on the same general approach: they train a DNN (or any DL-based approach), on only normal data (the rule of semi-supervised methods), to perform a specific task. Hence, they produce the desired result for the normal test data (that is, they detect no anomaly), since they have been previously seen during the training. The results would not be as desired for abnormal test data (that is, they detect an anomaly). The main challenge is to specify the desired scope of the task and select training and testing data accordingly. This way,  DNNs would learn and use proper features of the video data and thus discriminate well between normals and anomalies at the test time. The following sections review important methods using this common approach, describing their strategies, their feature extraction procedure, and their strengths and weaknesses.

\subsection{Reconstruction-based methods}
In reconstruction based AD methods, it is assumed that the models, trained by normal data, are able to reconstruct the normal test data accurately (i.e., with low reconstruction error), while the reconstruction error would be comparatively high for abnormal test data, which has not been observed by the model, during the training \cite{b37}. This methodology can be implemented in different ways (especially, using DNNs). In various research studies, deep Auto-encoder networks (especially Conv-AE) have been used to learn to reconstruct normal data. AEs perform well in reconstruction of the data, on which they have been trained. They encode the input visual data (a single frame, or a sequence of frames) into the latent space through an encoder and reconstruct the input data through a decoding pathway. For anomalies (the data samples, not seen in training), it is expected that the reconstruction error would be comparatively high \cite{b37}. An anomaly score (or vice versa, a regularity score) is normally calculated from the reconstruction error to indicate the anomalies. Figure 6 illustrates the process of reconstruction-based video anomaly detection. 

\begin{figure}[htp]
    \centering
    \includegraphics[width=8cm]{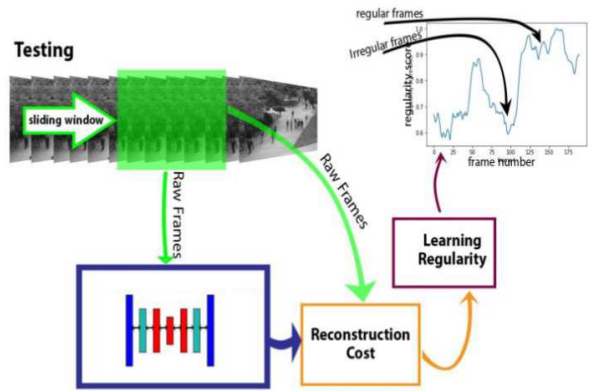}
    \caption{Reconstruction-based video anomaly detection, using AEs. This figure originally appeared in \cite{b36.1}.}
    \label{fig:reconstruction}
\end{figure}

One of the first and noticeable works in this field is proposed by Hasan et al. \cite{b37}, which uses a Conv-AE to extract spatiotemporal features from video clips and calculate the anomaly score from the reconstruction error. Although it uses a group of consecutive frames as input, instead of a single frame, to enforce the model to capture temporal dependencies, the 2D convolution destroys the temporal information, after the first convolution layer \cite{b37.1}. This issue has been addressed in \cite{b15}, which proposes a Conv-LSTM-AE to learn spatiotemporal features. The model extracts the spatial features of the frames, by a Conv-encoder, and passes them to a LSTM encoder to track temporal variations; the output goes through a reverse (temporal and spatial) decoder to reconstruct the frames group. Conv-LSTM-AE has also been used in \cite{b38} for anomaly detection.  In other similar works such as \cite{b30} and \cite{b39} a similar approach has been used for anomaly detection; however, they model normal data by minimizing the difference between the latent spaces of the input frame and the reconstructed frame, in addition to minimizing the reconstruction error of the frame itself. The work in \cite{b40} proposed to reconstruct the optical flow map of each frame, in order to consider the motion and to detect the anomalies in video. Moreover, some other researchers have proposed the concatenation of the appearance (frame) and motion data (optical flow), as an input, for the purpose of reconstruction. Nguyen and Meunier \cite{b27} proposed to use two different branches for motion and appearance, in order to capture the correspondence between them and to detect the anomalies more effectively. In this way, one branch is in charge of frame reconstruction (capturing spatial dependency) and the other one attempts to estimate the optical flow map, to capture motion dependency, customized for the task. Different models that have been applied for representation learning or reconstruction-based anomaly detection are as follows: PCA \cite{b40.1}, classic AE \cite{b37}, Conv-AE \cite{b37}, Contractive-AE \cite{b41}, Conv-LSTM-AE \cite{b15,b19}, Hybrid Spatio-Temporal Autoencoder \cite{b41.1}, Denoising AEs \cite{b42} and VAE \cite{b43}, GRU-AE \cite{b44}. Some of the other examples in this field are \cite{b35.3,b45,b46,b47,b48}. Manassés et al. \cite{b50} study the deep convolutional auto-encoders for anomaly detection in videos.

\subsection{Challenges of Auto-encoders for anomaly detection}
The challenges and shortcomings of Auto-encoders in anomaly detection are:\\
\\
1: AEs have a high learning capacity and a good power of generalization. Hence, the assumption that anomalies have a high reconstruction error is not always true \cite{b3, b50.1}.\\
2: When an AE is trained to minimize the Mean Square Error (MSE) of frame reconstruction, the network actually learns the average of previously seen training data.\\
3: Anomalies occurring in small regions can be neglected, because of the adding and averaging process for the entire frame, which may produce a low reconstruction error for anomalies in small regions \cite{b27}.

\subsection{Generative models for reconstruction}

Generative Adversarial Networks (GANs) and Variational AutoEncoders (VAEs) are also widely used in reconstruction-based anomaly detection methods. The main difference between these methods and previously introduced approaches is that these methods consider the distribution similarity, in addition to pixel-wise similarity (conventional reconstruction cost in AEs). Some of the noticeable AD works, based on GANs, include the researches conducted by Zenati et al. \cite{b51}, Kimura et al. \cite{b52}, Akcay et al. \cite{b53}, Akcay et al. \cite{b53.1}, Sabokrou et al. \cite{b54}, Gherbi et al. \cite{b55}] and Ganokratanaa et al. \cite{b56}. Zenati et al. \cite{b57} and Donahue et al. \cite{b58} use a biGAN to map a latent space to an image and use it for anomaly detection. Gans have promoted the performance for various AD approaches, especially prediction-based approaches. However, as GANs may show instability during training, their usage for anomaly detection may be limited. Hence, in order to address this problem, in addition to comparing frames, extracted features are also compared to calculate the loss \cite{b53}. Galeone et al. \cite{b59}, Rani and Sumathi \cite{b59.1} have, comprehensively, studied GANs for anomaly detection. Figure 7 shows 3 state-of-the-art GAN-based AD architectures.

\begin{figure*}[htp]
    \centering
    \includegraphics[width=11cm]{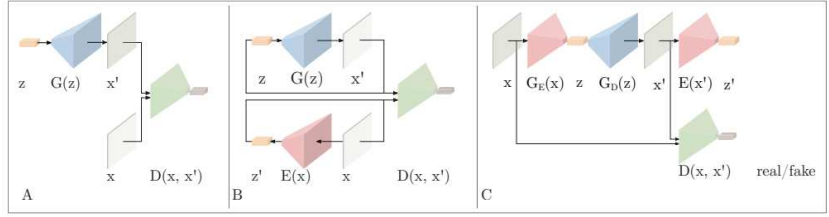}
    \caption{Different GAN-based AD methods \cite{b53}. A: AnoGAN \cite{b64}, B:Efficient-GAN-Anomaly \cite{b57}. C: GANomaly \cite{b53}.}
    \label{fig:GAN}
\end{figure*}

\subsection{Prediction-based methods}

Prediction, here, means the estimation of future frame(s), based on previously seen frames. In prediction-based anomaly detection methods, it is assumed that predictive models, which are trained on normal sequences (previously seen frames), can precisely predict the future frame(s) in normal test sequences but their prediction error would be comparatively high in abnormal test sequences. Thus, in video anomaly detection, video frames are considered as sequences and the goal is learning the normal patterns (appearance and motion), in consecutive normal frames and predicting the coming frame, based on the learned patterns. The prediction error can be easily calculated by measuring the difference between real and predicted frames or by calculating the conditional probability of a new observation based on the previous samples \cite{b9}. Different constraints have been used for anomaly detection, in prediction-based frameworks, such as appearance (gradient and intensity) and motion \cite{b3}. Experiments in \cite{b3} show that predicted frames for abnormal samples are unclear and usually with color distortion and it is claimed that among several networks, GANs show better results for video prediction. Wang et al. \cite{b60} report that feature extraction, through the prediction process, has high quality and it is more suitable for video analyzing applications, since accurate prediction highly depends on high quality features. It is worth mentioning that, in prediction-based methods, the input and output are not necessarily of the same type or size and they can be different, in different approaches. For example, \cite{b60.01} takes advantage of two cross-domain generators, in which one learns predicting the past gradients from appearance and the other learns the reverse, for local anomaly detection. Prediction-based video anomaly detection strategy has been utilized in numerous research studies such as: \cite{b3,b61,b62,b63}.

\subsection{Prediction versus Reconstruction}

A prediction-based method attempts to obtain the most information from the most recent frames, as they are more relevant to the future frame \cite{b19}. Hence, predictive methods lose a lot of information about the past and their generic (general) prediction would be less precise. Moreover, Pathak et al. \cite{b36.01} declare that as nearby frames are visually similar (considering the texture and the color), they might focus on learning low level features instead of high level semantic features. Reconstruction, on the other hand, attempts to learn an obvious representation from data \cite{b19} and in fact, it memorizes the input \cite{b9} and considers all frames almost equally. In this way, it neglects the temporal evaluation between frames. To address the mentioned challenges, a composite approach has been proposed to benefit from the advantages of both methods \cite{b44,b65,b66}. The proposed LSTM-AE networks are composed of two branches, one for reconstruction and one for prediction. These branches have an encoder in common but with two separate decoders. One of the challenging aspects of both prediction based or reconstruction based methods is that even slight lighting variations may cause a high pixel-based loss, which can be deceptive. Moreover, these approaches generally train the model (reconstruction or prediction) from scratch, in an unsupervised manner, and the entire frame or only proposal patches are reconstructed or predicted. Hence, these approaches are not aware of the class of the objects in the frames, or in the proposal patches. To address these challenges,researchers such as Bergmann et al. \cite{b67} use a pre-trained network (trained on natural images) as the encoder. Producing the latent space in this way, helps the network leverage prior knowledge about the nature of the natural images and tackle the issue to some extent.

\begin{figure}[htp]
    \centering
    \includegraphics[width=7cm]{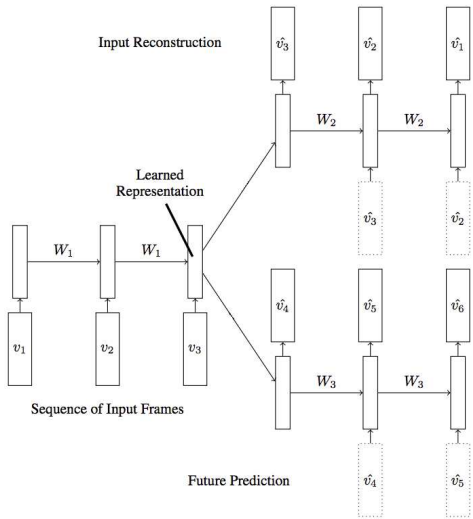}
    \caption{Combination of frame reconstruction and prediction for anomaly detection in video \cite{b65}.}
    \label{fig:GAN}
\end{figure}

\subsection{Object-centric based methods}

As mentioned before, one of the main shortcomings of the methods based on frame reconstruction or prediction is that they do not explicitly (and hence effectively) consider the objects. Object centric approaches concentrate on detected objects (detected by state-of-the-art object detectors) and study their appearance and motion features to make decisions. The researach conducted by Ionescu et al. \cite{b68} is one of the recent works on video anomaly detection that detects objects of interest to accomplish anomaly detection. Moreover, Doshi and Yilmaz \cite{b69} propose an object centric approach, in which objects of interest in each frame are detected by a pre-trained YOLOv3 object detector and consequently, a feature vector containing appearance, motion, and location information is extracted to learn normal behaviors. Unlike Ionescu et al. \cite{b68}, this method considers location  information by containing a summary of location information in its provided feature vector. Other researchers have also detected anomalies by detection of objects \cite{b70,b71,b72,b73}. The advantages and challenges of object-centric based methods are:\\
\\
- The objects are explicitly considered, which is helpful in video understanding.\\
- The anomalies are easily located inside the frame.\\
- These methods extract and consider short term motion information.\\
- The performance of the method completely relies on the object detection part.\\
- Information regarding the object size and he context (such as the location information) is removed as these methods crop and resize the detected objects.

\begin{figure}[htp]
    \centering
    \includegraphics[width=8cm]{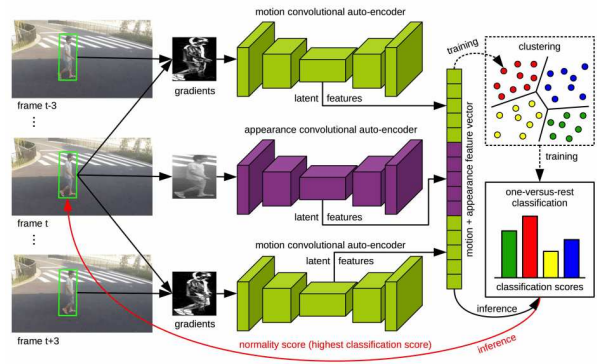}
    \caption{An object-centric video anomaly detection method, proposed in \cite{b68}.}
    \label{fig:object centric}
\end{figure}

\subsection{Segmentation-based methods}
Krzysztof et al. \cite{b74} perform the anomaly detection task in a different way. The proposed idea arises from the fact that a semantic segmentation approach can segment the objects properly, if it has observed them in the training phase and it would show worse results for unseen objects. This fact can be used for image anomaly detection. The researchers propose to synthesize the image from produced semantic segmentation maps and the reconstructed images help to define and locate the novel objects. The positive point of this method is object type awareness, however this method is proposed for images, not videos. Another research \cite{b75} proposes a similar anomaly detection approach, based on foreground segmentation and detects the unexpected objects (i.e., objects not seen in the training samples). However, this work also does not take actions and events into account.

\begin{figure}[htp]
    \centering
    \includegraphics[width=8cm]{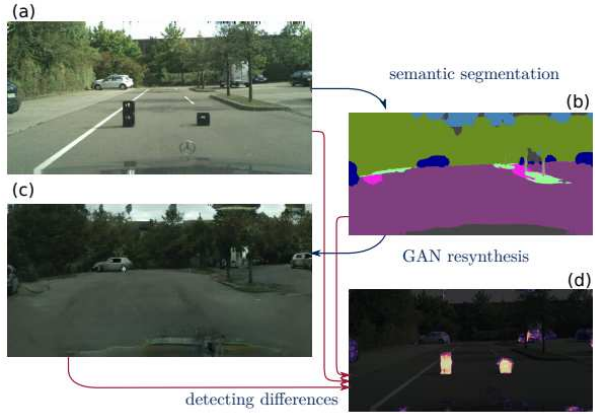}
    \caption{Anomaly detection, based on semantic segmentation \cite{b74}. a: input frame. b: extracted semantic segmentation map. c: resynthesized frame from segmentation map. d: difference of input and resynthesized frame. The image in this example comes from the Lost and Found dataset \cite{b75.1}.}
    \label{fig:segm}
\end{figure}

\subsection{Memorization-based methods}

One of the main challenges with previous methods is that DNNs (and especially CNNs) are so powerful in generalization, that they may reconstruct the abnormal frames too well. Hence, the assumption that the reconstruction error is comparatively high for abnormal test frames is not always true. In order to address this problem and reduce the representation power of DNNs, memorization-based anomaly detection methods have been proposed. These methods use the encoding of the input frame as a query to select the most relevant saved items, from the recorded prototypical patterns of normal data, to reconstruct the input frame. Consequently, the previously recorded items are decoded and selected from memory, instead of using the output of the encoder directly. For example, Gong et al. \cite{b76} proposed MemAE (Memory augmented AutoEncoder) which learns and updates the memory contents, during training, to represent the prototypical elements of the normal data. In the test phase, the memory is fixed and reconstruction is performed using items selected from the memory (see Figure 11 and 12). Moreover, Park et al. \cite{b77} propose a similar strategy for anomaly detection and reconstruct or predict a video frame with a combination of items in the memory, rather than using CNN features directly from an encoder. In this work, items in the memory record prototypical patterns of normal data and the diversity of normal patterns is considered explicitly, since the authors believe that a single prototypical feature is not enough to represent various patterns of normal data.

\begin{figure*}[htp]
    \centering
    \includegraphics[width=7cm]{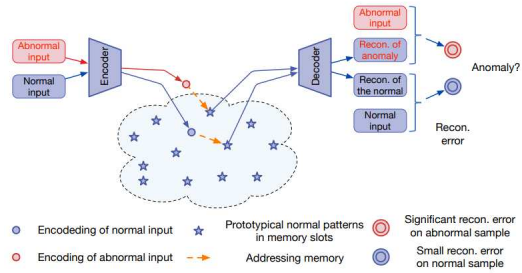}
    \caption{Illustration of the memorization-based anomaly detection \cite{b76}.}
    \label{fig:mem}
\end{figure*}

\begin{figure*}[htp]
    \centering
    \includegraphics[width=7cm]{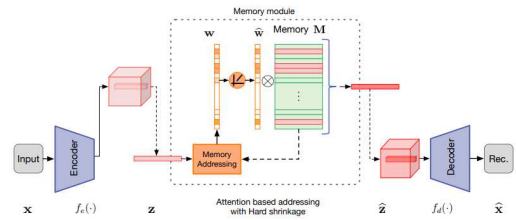}
    \caption{Diagram of the proposed MemAE \cite{b76}.}
    \label{fig:memae}
\end{figure*}

\begin{table*}[!hbt]
 \caption{Comparison of different DL-based AD strategies from different viewpoints. S and ST stand for Spatial and spatiotemporal respectively.}

{\footnotesize
 \begin{adjustwidth}{-0.45cm}{}
\begin{tabular}{|p{4.5cm} ||p{2.2cm}| p{1.8cm}|p{1.9cm}| p{2.3cm}| p{2.1cm}| }

\hline
\textbf{Strategy used for AD} & \textbf{Reconstruction} & \textbf{Prediction} & \textbf{Segmentation}& \textbf{Object Centric} & \textbf{Memorization }\\
\hline
\hline

\textbf{Object class awareness} & No & No & Yes & No & No \\
\hline

\textbf{Generalization of model to anomalies} & Yes & Yes & Low & Yes & No \\
\hline

\textbf{Extracted features} & ST & ST & S\footnote{1} & ST\footnote{1}  & ST \\
\hline

\textbf{Used DNNs} & AE, Conv-AE & Conv-LSTM-AE, GANs,Unet & GANs & AE, Conv-AE & AE, Conv-AE \\
\hline

\textbf{Aware of environment and contextual information (location, time)} & Implicitly  & Implicitly  & No & No & Implicitly \\
\hline 

\textbf{Some examples} & \cite{b15,b37,b40} & \cite{b3,b7,b60} & \cite{b74} & \cite{b68,b70,b71,b72,b73} & \cite{b76,b77} \\
\hline

\end{tabular}

 \end{adjustwidth}
 }

\end{table*}

\subsection{Shortcomings or challenges of previous methods}
Various DL-based semi-supervised video anomaly detection methods were critically analysed in the previous sections. Based on their experiments and conclusions, some challenges and shortcomings can be summarized as below:\\
\\
1: Previous reconstruction/prediction based methods usually reconstruct/predict the entire frame and consider the frames and the motion with a global look. They do not consider the objects and other details individually. Some other methods only attempt to focus at reconstruct or prediction of the objects, and therefore fail to consider the context. Hence, almost all existing methods neglect some important information in their algorithm.\\
2: A considerable portion of spatiotemporal information in frames is redundant, and is not required in scene analyzing or video understanding. This leads the network to divide its attention to a variety of aspects (including these redundant parts) and not to precisely focus on useful portions (for example, to the objects of interest). This fact plays an important role in video anomaly detection, since anomalies generally occur rarely and may occupy a small portion of a frame. This problem has not been acceptably covered in existing methods.\\
3: Loss functions, which direct the network to capture effective features, do not simultaneously and effectively apply compactness and descriptiveness constraints to the feature extraction process.\\
4: Existing methods do not effectively take the class of the objects into account. Moreover, the relation between class of the object of interest, its motion, and location has not been taken into account effectively.\\
5: In the existing methods, if an anomaly occupies a small portion of the frame, its effect could be lost on the anomaly score of the frame. On the other hand, even object-centric methods may also have a difficult time detecting such small objects.\\
6: The currently used approach for the calculation of the reconstruction error is not reliable. Any small changes in all pixels (for example illumination changes in the environment) can result in a high change in reconstruction error.\\
7: Holistic models, trained on a scene, may not perform well after a change in the scene or view point.\\
8: In existing methods, the fusion of different anomaly scores (e.g., motion anomaly score, appearance anomaly score, etc.) do not apply the effect of all factors effectively and one factor may dominate others and lead to underestimation of other factors.\\
9: DNNs trained on normals may sometimes be generalized too well on anomalies.

\section{Experiments and results}

In the previous sections, critical review and analysis of different methods were provided based on the results of previous research studies. These analyses would also be concluded logically, considering different factors such as the architecture of the proposed network, target function, considering results of similar applications, etc. This section provides the results of experiments that we conducted to clarify some of the points which have not been considered and discussed in the previous works. These points can be listed as follows: effect of the number of the foreground objects, effect of the camera distance (or the object size), awareness of the method concerning the class of the objects, effect of motion patterns and illumination changes. To analyze these points, two state-of-the-art methods were implemented, from which we draw conclusions for these methods and also for other methods which have similar parts or steps in common. These experiments were conducted from a different point of view: examining the failures of methods in some frames. In other words, instead of comparing the performance of different methods by numbers, we attempt to highlight the strengths and weaknesses of different methods so that they can be considered and addressed in future works in order to reduce false positives and false negatives.\\

First, we implemented the method proposed by Hasan et al. \cite{b37} which uses a Conv-Autoencoder for anomaly detection. It is worth mentioning that Hasan et al. conducted experiments on two different autoencoder architectures: Fully Connected Autoencoder (FC-AE) and Convolutional Autoencoder (Conv-AE). In addition to the experimental results that they provided, it also can be concluded (considering the network architecture) that FC-AE destroys the structure of the image and it could not show comparatively promising results for image processing. Hence, we do not implement that part. To train the Conv-AE, we implemented the same architecture and used the same hyperparameters as originally proposed (The implemented architecture is shown in Figure 13).\\

\begin{figure*}[h]
    \centering
    \includegraphics[width=8cm]{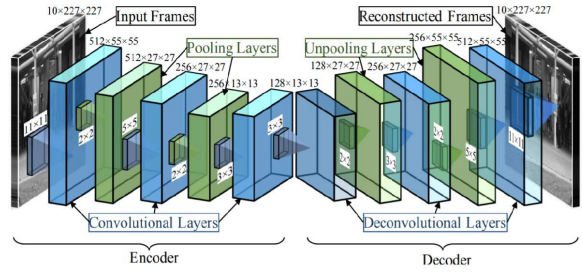}
    \caption{The architecture of the Conv-Autoencoder proposed in \cite{b37} for video anomaly detection. }
    \label{fig:hasan}
\end{figure*}

After training the model on the UCSD dataset, we performed an evaluation on its test dataset. We calculated the reconstruction error and consequently the regularity score for each frame as in equations 1 and 2. In these equations, s(t) and e(t) show the regularity score and reconstruction error of the frame, respectively. I(x,y,t), e(x,y,t) also refers to the intensity and the reconstruction error of the pixel.\\

\begin{equation}
    e(x,y,t)= \| {I(x,y,t)-f_{w}(I(x,y,t))} \|^2
\end{equation}

\begin{equation}
    S(t)= 1-\frac{e(t)-min_{t}e(t)}{max_{t}e(t)}
\end{equation}\\

As can be seen in Figure 14, the results of the experiments show that this method fails when the number of foreground objects is considerably variable in different frames. As Figure 14 shows, when the number of the foreground objects is high, we would have a lower regularity score (or higher reconstruction error) for the frame. This is because each object, more or less, has a reconstruction error and as the number of the objects rises, the total reconstruction error of the frame, which is the sum of the errors of the foreground objects and the background (BG), increases. From another point of view, as Figure 15 shows (This figure originally appeared in \cite{b37}), the most regular frame for each scene is an image quite similar to its BG. BG pixels are the constant and the most frequent pixels in all images during training and the network easily learns them. The reconstruction error of the frame is due to the difference between the input frame and the most regular frame (let us assume BG here), and is thus directly affected by the number of objects. It can be concluded that, for cases in which the class of the objects defines the anomalies rather than their number, objects should be analyzed individually (as with object-centric approaches) instead of evaluating the entire frame at once. We also repeated the same analysis for the training samples (Figure 16) and the experiment confirms the previous results. That means that the reconstruction error of the frames with high populations (even if they do not contain any anomaly) is considerably higher than that of the other frames (even compared to frames with anomalies).\\

\begin{figure*}[h]
    \centering
    \includegraphics[width=8.5cm]{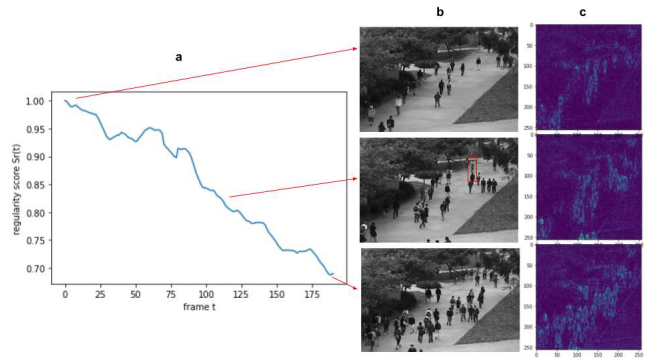}
    \caption{Effect of the number of the FG objects on the regularity score (UCSD-Ped1-Test003). (a) Regularity score as a function of frame number. (b) Samples of frames in the test clip. (c) Reconstruction error map for the input frame (e(x,y,t)). }
    \label{fig:comp}
\end{figure*}

\begin{figure*}[h]
    \centering
    \includegraphics[width=8cm]{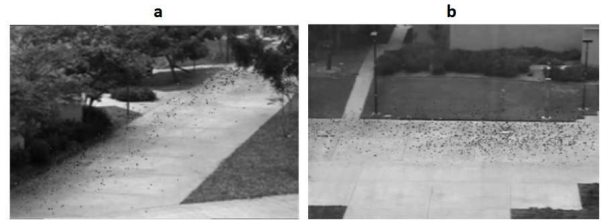}
    \caption{(a) Synthesized regular frame for Ped1. (b) Synthesized regular frame for Ped2.}
    \label{fig:regular}
\end{figure*}

\begin{figure*}[h]
    \centering
    \includegraphics[width=9cm]{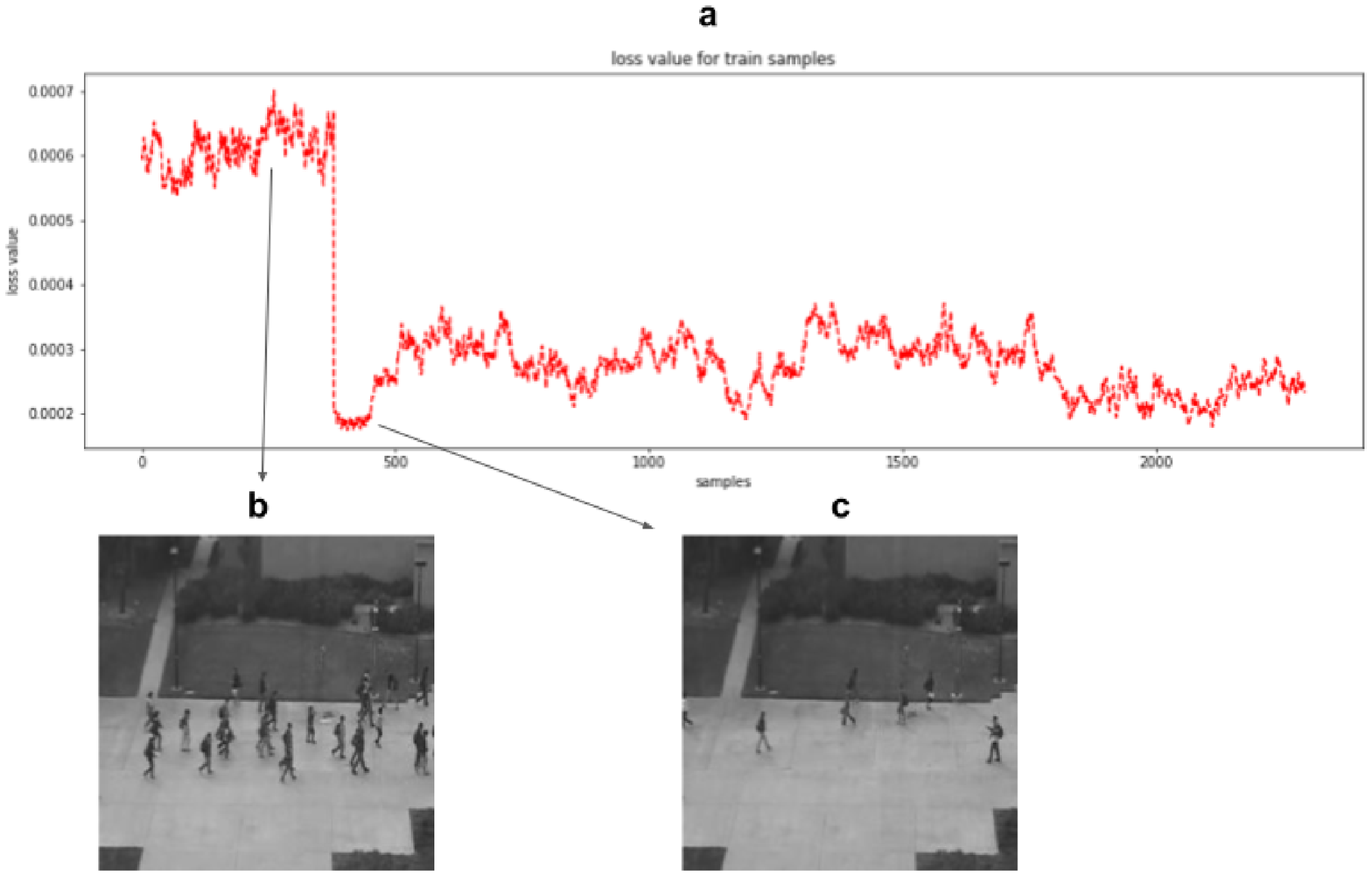}
    \caption{Effect of the number of the foreground objects on the reconstruction error (results on Ped2). (a) Reconstruction error (loss value) for the training frames. (b) A frame with a high number of people. (c) A frame with a low number of people.}
    \label{fig:train}
\end{figure*}

In most of the recent research studies, the appearance and motion features were analyzed separately, in separate branches. In the motion branch, researchers reconstruct or predict the previously extracted motion features (such as optical flow or difference of consecutive frames) to learn the normal motion patterns. We analyzed the effect of the number of objects on the results of the motion branch. As can be seen in Figure 17, the results give rise to the previous conclusion; if the frame (or the motion map) is analyzed globally (analyzing the entire frame, not each object) the reconstruction error would be affected by the number of foreground objects, rather than the important anomaly factors.\\

\begin{figure*}[h]
    \centering
    \includegraphics[width=9cm]{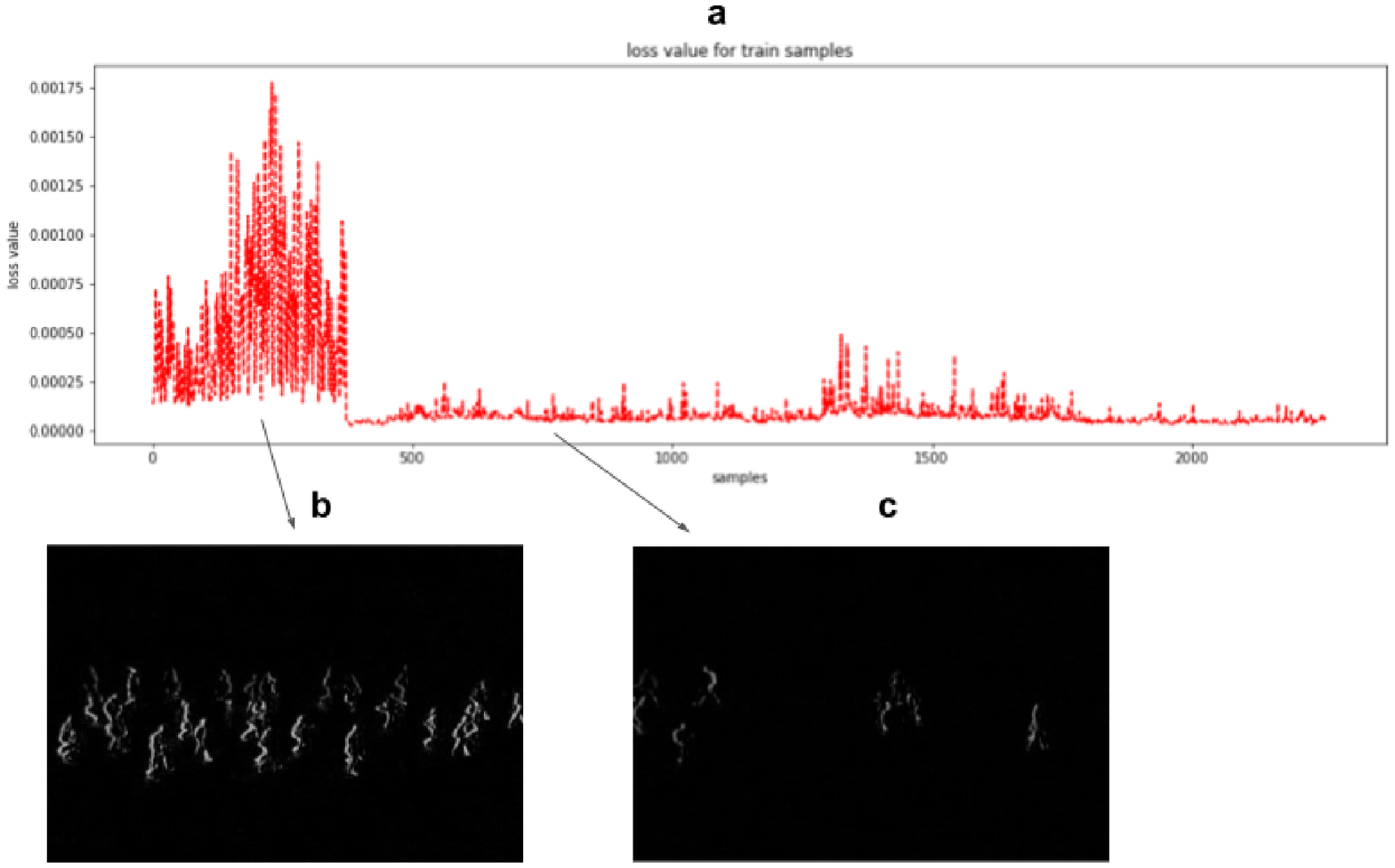}
    \caption{Reconstruction error for motion maps. (a) Reconstruction error (loss) of motion maps for training samples. (b) Motion map of a frame with a high number of people. (c) Motion map of a frame with a low number of people.}
    \label{fig:motion}
\end{figure*}

We should make this point clear that the mentioned shortcoming has nothing to do with formulating the anomaly detection as a reconstruction or a prediction problem but is due to the fact that these methods consider the frame holistically. As object-centric approaches analyze each object individually, they do not face this challenge.\\

From another point of view, considering the frame entirely has a strong point: Considering location. Frame reconstruction or prediction-based methods, mostly learn a pixel-wise model. This means that they learn a model for each pixel separately, and pixels at different locations expect different intensities. Before examining the experiment results, let us have a second look at Figure 15-b, for example. Figure 15-b shows that the most regular frame for the Ped2 has many dark pixels in the upper side of the walkway. This is because during training, the network has frequently seen dark objects in that side. From this image it is more expected (i.e., it is normal) to see dark objects in the upper side and the presence of dark objects in the lower part of the walkway most probably would be detected as an anomaly. To validate this idea, we analyzed the response of the network in some frames of Ped2, which is shown in Figure 18. As expected, the network considers the position of the object in the scene and it produces higher error for objects which are not expected in that location. This positive feature is missed in object-centric approaches, as they crop objects out of the frame and analyze them individually. Thus, the location information is either lost or not considered. This would produce false positives and false negatives for object-centric approaches in different datasets, especially the street scene dataset. As can be seen in Figure 1, in the Street scene dataset, the definition of normality (and hence anomaly) is different for 4 different cars, considering their locations.\\

\begin{figure*}[h]
    \centering
    \includegraphics[width=6cm]{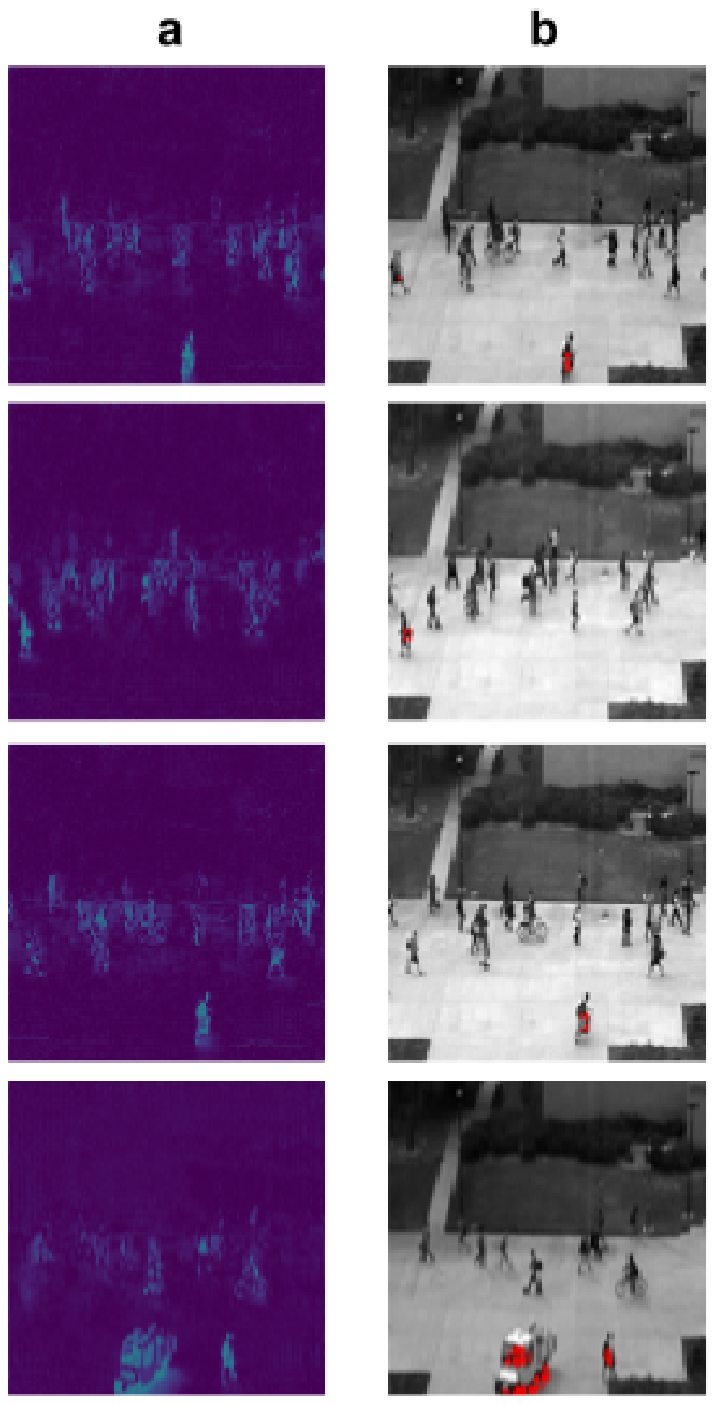}
    \caption{Different response of the network to the same pixel intensities at different locations. (a) Reconstruction error map (b) Input Frame with detected anomalies in red.}
    \label{fig:location}
\end{figure*}

Figure 19 shows the results of anomaly detection for different sample frames. This figure shows the input image (a), reconstructed image by the model (b), reconstruction error map or the differences between original and reconstructed image (diff) in (c), and finally (d) indicates the pixels which are most probably anomalies, by applying a threshold on the reconstruction error map. This figure highlights a few points: First, what is missing in existing methods (in most reconstruction and prediction-based methods) is that the effect of the distance of the object from the camera is not considered. As illustrated in the figure, as the car approaches the camera, it occupies many pixels in the frame and produces a higher reconstruction error for the entire frame. Hence, the anomaly score of the frame would be affected by the factor of object distance. Another point that can be concluded from the results is that these methods mostly consider the intensity (or color) of the objects instead of the class of the objects. As illustrated in the results, the anomaly points (red points) are only detected for the pixels which have an intensity which differs from that of the background and the other parts of the car are detected as normal. This can also be concluded logically; in these methods, the model is trained to reduce the Mean Square Error of the pixel’s intensity (low-level features), which causes it to focus on low-level features. On the other hand, no information regarding the class of the objects is provided to the model directly.\\

\begin{figure*}[h]
    \centering
    \includegraphics[width=8cm]{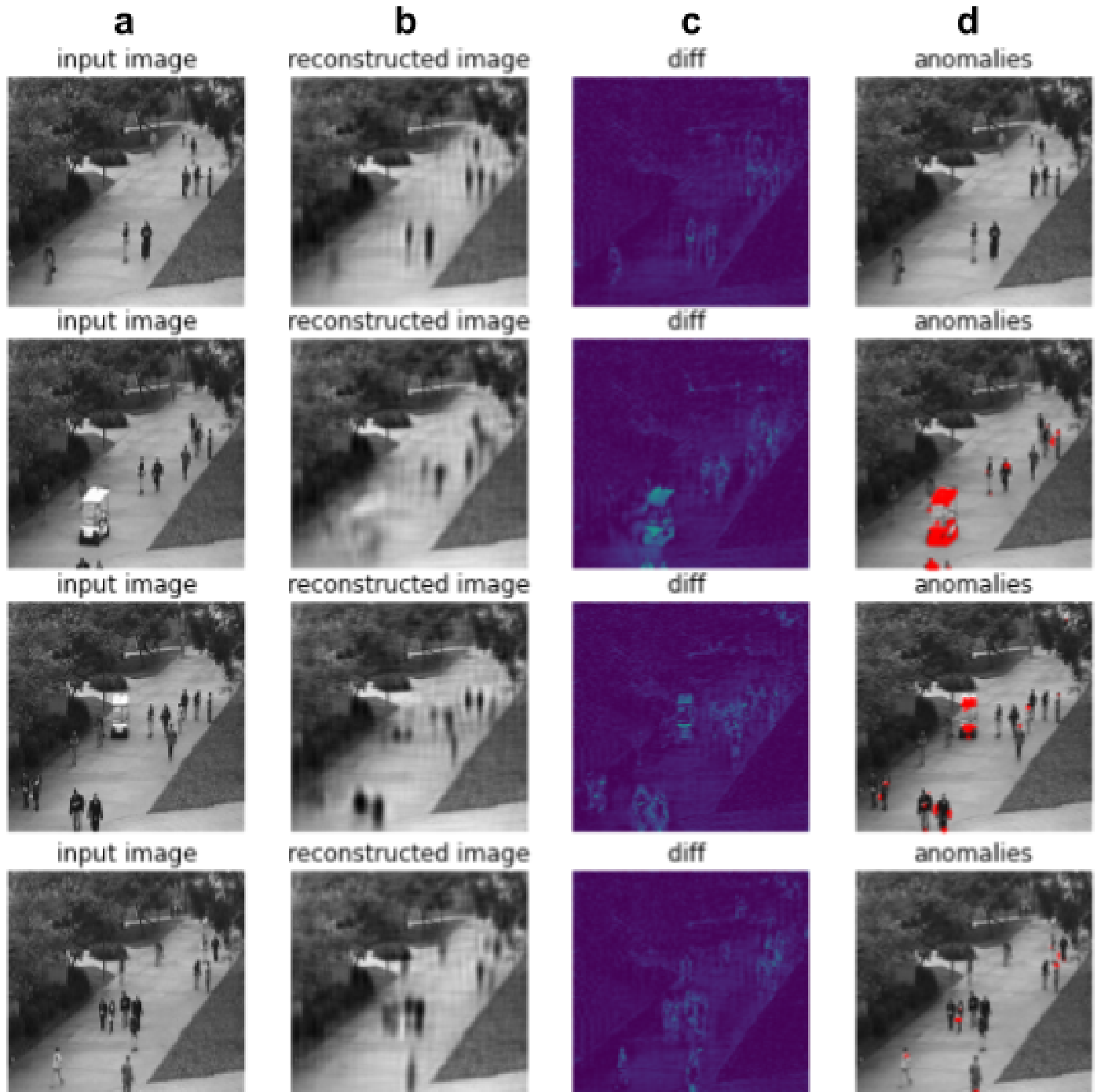}
    \caption{Effect of the camera distance and pixel intensity on anomaly detection (results are provided for UCSD-Ped1). (a) Input frame. (b) Reconstructed frame. (c) Difference between input and reconstructed frame (reconstruction error map). (d) Result of anomaly detection (anomalous pixels are indicated in red).}
    \label{fig:rec}
\end{figure*}

The method proposed by Hasan et al. does not effectively consider motion, because as mentioned in section 3.3, the first convolutional layer destroys the temporal information. Furthermore, our experiments do not show any considerable reconstruction error for objects with abnormal motion (faster motion here) such as bikers and skateboarders. Although in the results there is a noticeable reconstruction error for the cars, due to their different pixel intensity and their comparatively larger size. Hence, in the second step, we implemented the method proposed by Chong et al. \cite{b15}. This method benefits from a temporal autoencoder which is embedded inside the spatial autoencoder. We implemented the same architecture, as proposed originally \cite{b15}. This architecture is shown in Figure 20.\\

\begin{figure*}[h]
    \centering
    \includegraphics[width=8cm]{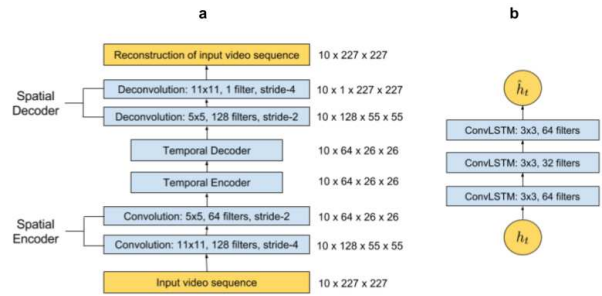}
    \caption{The proposed Conv-Lstm-Autoencoder in \cite{b15}. (a) The entire architecture. (b) The temporal autoencoder.}
    \label{fig:conv-lstm}
\end{figure*}

The results indicate that the previous challenges in Conv-AE such as the effects of distance and number of objects, unawareness regarding the class of objects, etc. still exist here, since this approach has similar strategies (such as evaluating the entire frame at once, focusing on intensity, etc.). However, as this method adds a temporal autoencoder to the network, it can capture motion patterns.\\

\begin{figure*}[h]
    \centering
    \includegraphics[width=8.5cm]{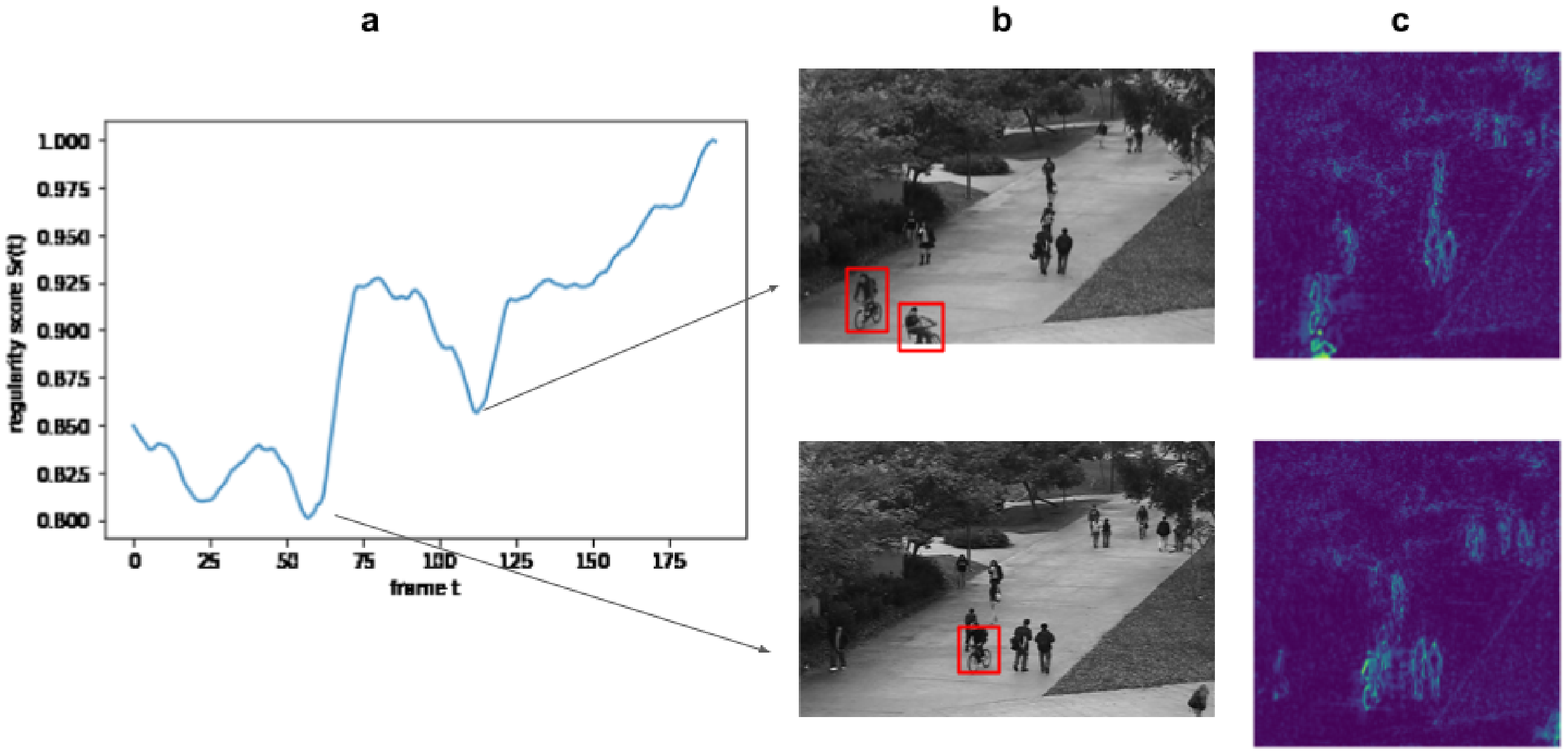}
    \caption{Results of Conv-Lstm-AE on Ped1 (Test033). (a) Regularity score of the clip. (b) Input frames. (c) Reconstruction error map for the input frames.}
    \label{fig:bikers}
\end{figure*}

As illustrated in Figure 21, bikers’ bodies (unlike other persons’ bodies) produce a higher intensity in error maps. The proposed method explicitly models the temporal evolution of the frames and hence can capture motion. However, the produced reconstruction error for the entire frame again depends on several factors which may degrade the effectiveness of the method. As it can be concluded from these results and also previous ones, these factors can be: 1) Number of foreground objects: the number of the objects is more decisive than the effect of the object motion. 2) Distance from the camera: the motion effect of an object, in the reconstruction error map, can be easily neglected if the object is located far from the camera (i.e., for smaller objects). 3) The results show that in single path methods (categories A and C in section 3.1), the effect of the appearance features may dominate the effect of motion features. It can be expected that two-branch approaches would produce better results in considering the effect of the motion anomalies, as they independently consider and analyze motion in a different branch. Figure 22 confirms the previous point. In this experiment, we removed 9 consecutive frames (frame 9 to 17) to synthetically generate a sudden motion between frames 8 and 20. Through these frames, all objects inside the scene are normal objects. In other words, we synthetically generated an abnormal motion for normal objects (i.e., the abnormality is simply due to the motion factor), and this motion is much faster than any motion in the clip. However, as Figure 22 indicates, the regularity score is not considerable.\\

\begin{figure*}[h]
    \centering
    \includegraphics[width=7cm]{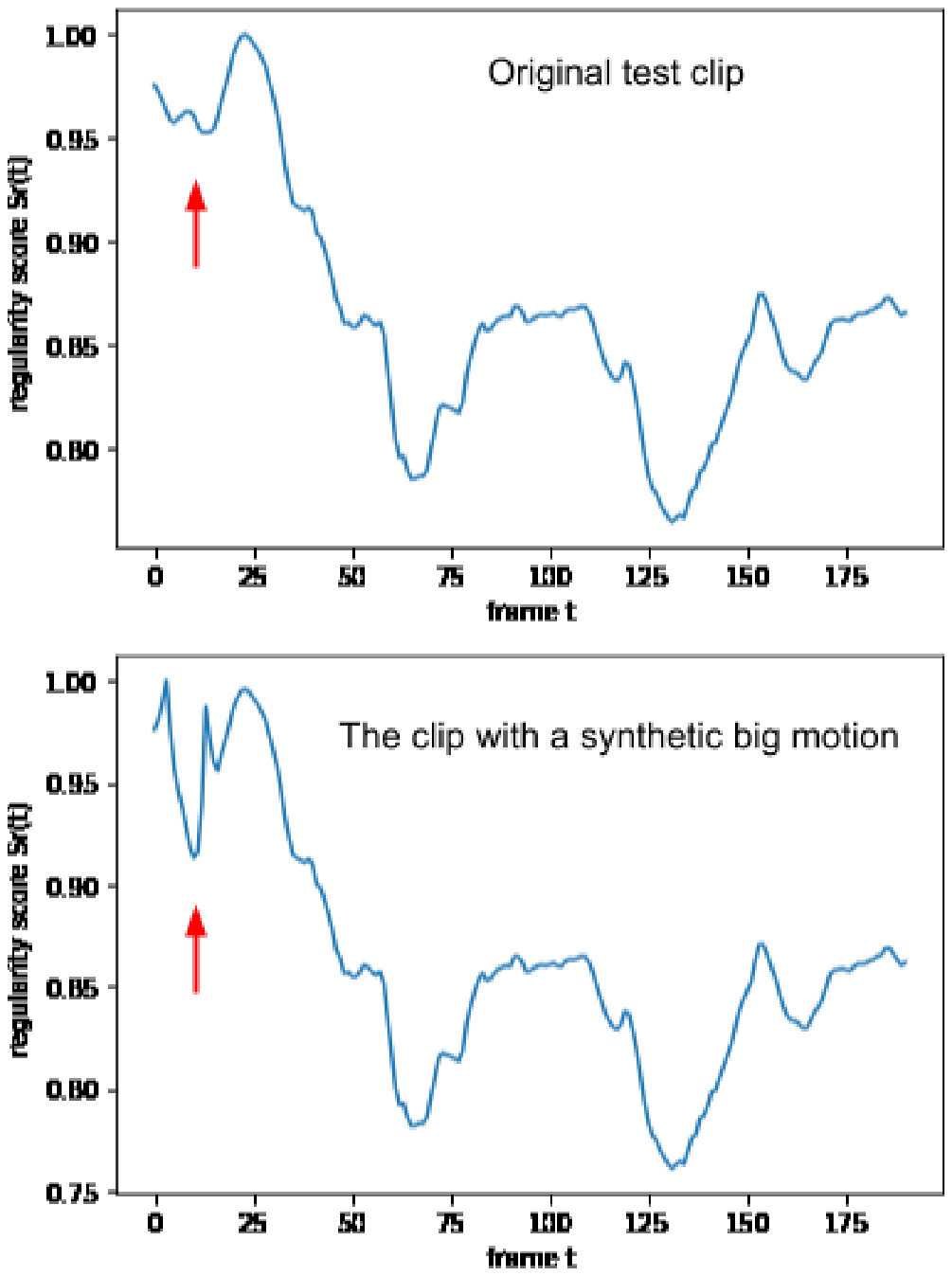}
    \caption{Regularity score for different motions. (a) Regularity score for an original test clip (ped1-test016). (b) Regularity score for the same clip with synthesized large motion.}
    \label{fig:bigmotion}
\end{figure*}

In another experiment, in order to analyze the effect of lighting changes in the scene we made changes in the brightness of some test frames. For this purpose, the pixel intensity of entire pixels in a test frame was multiplied to 1.3, and the reconstruction error map was extracted for the original test frame and the newly generated frame. The results, shown in Figure 23, indicate that by lighting changes in the frame, the reconstruction error map varies considerably, which effects the performance of the method. In other words, this not only shows that the system is vulnerable to lighting changes, but also that the system considers low-level features instead of focusing on high-level ones.\\

\begin{figure*}[h]
    \centering
    \includegraphics[width=7cm]{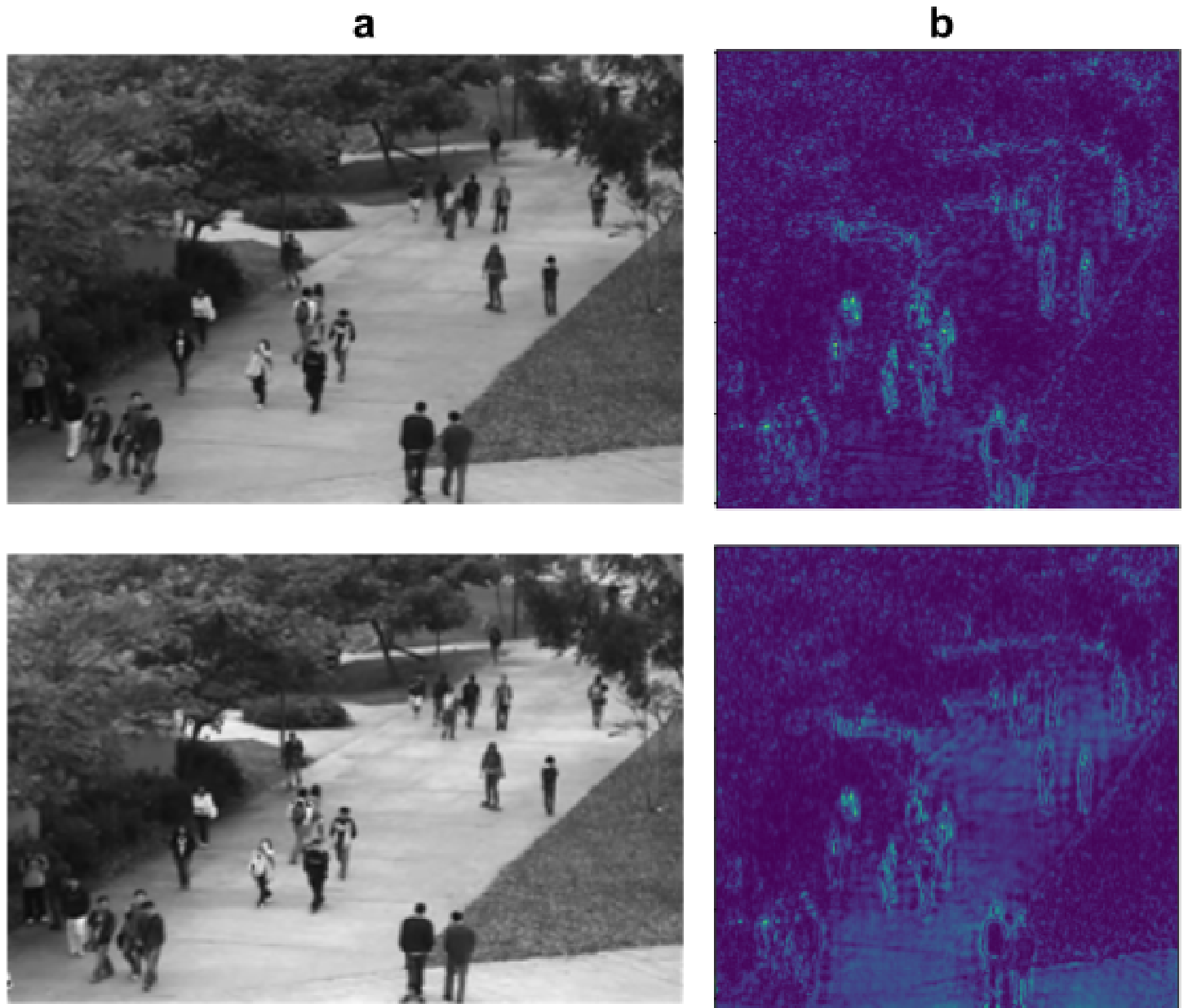}
    \caption{Effect of lighting changes on the reconstruction error map. (a) Input frames. top: original frame, bottom: the same frame after increasing the pixels intensities. (b) Reconstruction error map.}
    \label{fig:lighting}
\end{figure*}

Experimental results in Figures 24 and 25 illustrate the dominance of some factors such as intensity and distance on the class of the objects (which is the reason for the definition of the anomaly here). In Figure 24, the top two frames are both normal frames, however the second one would most probably be detected as an anomaly. In Figure 25, the frame in row 4 contains an anomaly in the far distance which results in producing a higher regularity score compared to row 1 which only contains normal objects.\\

\begin{figure*}[h]
    \centering
    \includegraphics[width=8.5cm]{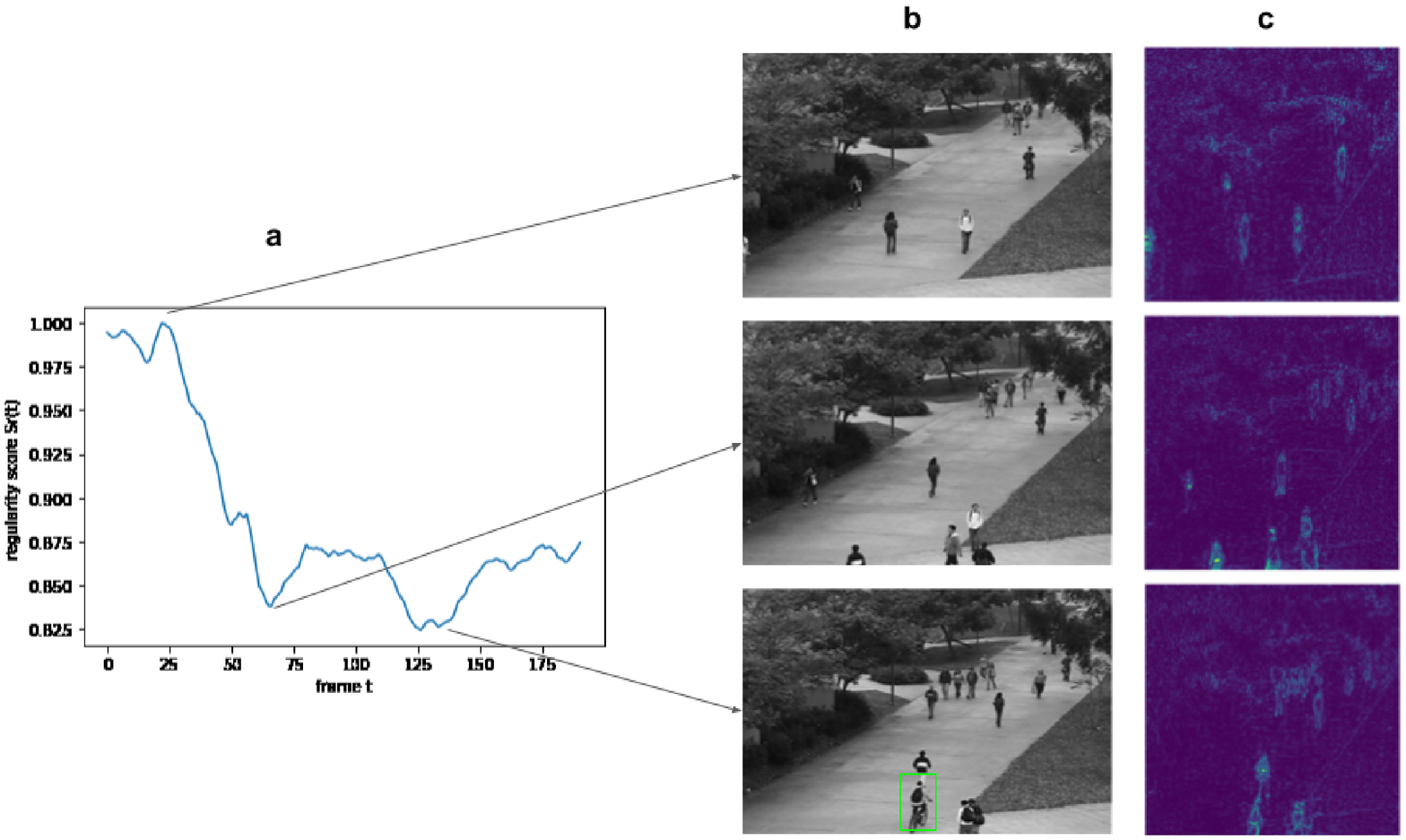}
    \caption{Results of Conv-Lstm-AE on Ped1 (clip 016). (a) Regularity score of the clip. (b) Input frames. (c) Reconstruction error map for the input frames.}
    \label{fig:3pic}
\end{figure*}

\begin{figure*}[h]
    \centering
    \includegraphics[width=8.5cm]{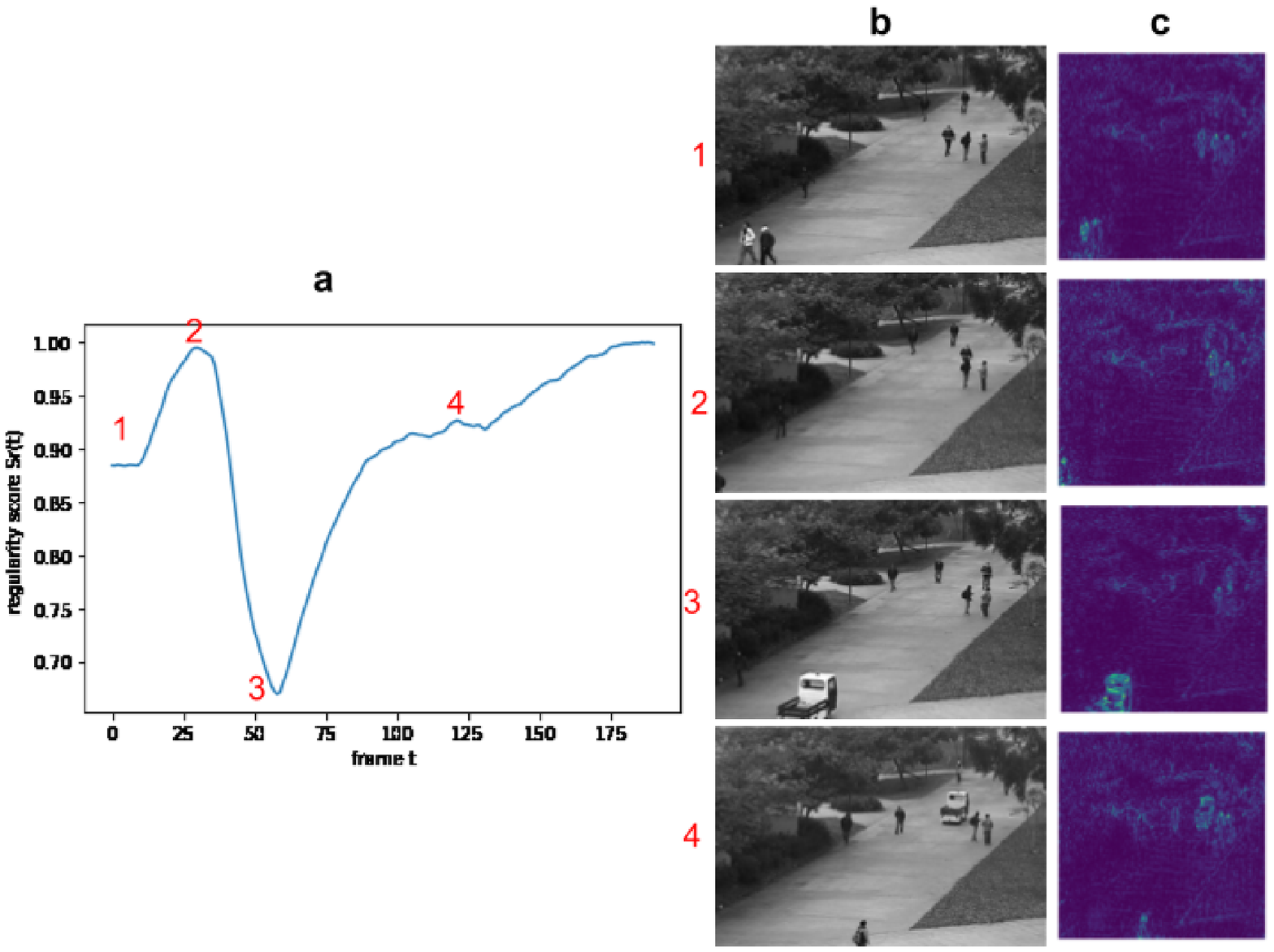}
    \caption{Results of Conv-Lstm-AE on Ped1 (clip 020). (a) Regularity score of the clip. (b) Input frames. (c) Reconstruction error map for the input frames.}
    \label{fig:4pic}
\end{figure*}

In the final step, we carried out a test to evaluate how effective object-centric approaches could be, in considering the class of the object and identifying abnormal objects. Object-centric approaches, as discussed in section 3.8 and as observed in Figure 9, crop objects out of the frame and train a network (usually an autoencoder) to learn normal patterns. However, as the experiments show, reconstructing a frame or even the appearance of objects individually, does not necessarily lead these approaches to consider the class of the object. This is mainly due to the fact that training an AE for the purpose of reconstruction, mostly focuses on the intensity (or color). What object-centric approaches mainly contribute, is focusing on the object, rather than other factors (such as BG or the number of objects, etc.). In order to validate this idea, we separately trained and evaluated two autoencoders with, respectively, cropped images of objects and their class-level features, which were extracted by a pre-trained CNN. For this experiment, two groups of images were prepared and named ‘Normals’ and ‘Abnormals’. The Normals group contains all of the cropped objects of the same group (here, images of people) and the Abnormals group contains cropped images of different types of objects (such as vehicles, bicycles, bikes, etc.). The Normals and Abnormals groups present normal and abnormal objects, respectively. Then, using a pre-trained VGG19, the class-level (i.e., features of the last layer), and color-level features (i.e., features of the first layer) were extracted for each group. Each network was trained and evaluated separately with images of normal objects and their class-level features. The evaluation was caried out on both normal and abnormal groups, the results of which are shown in Figure 26.\\

\begin{figure*}[h]
    \centering
    \includegraphics[width=8.5cm]{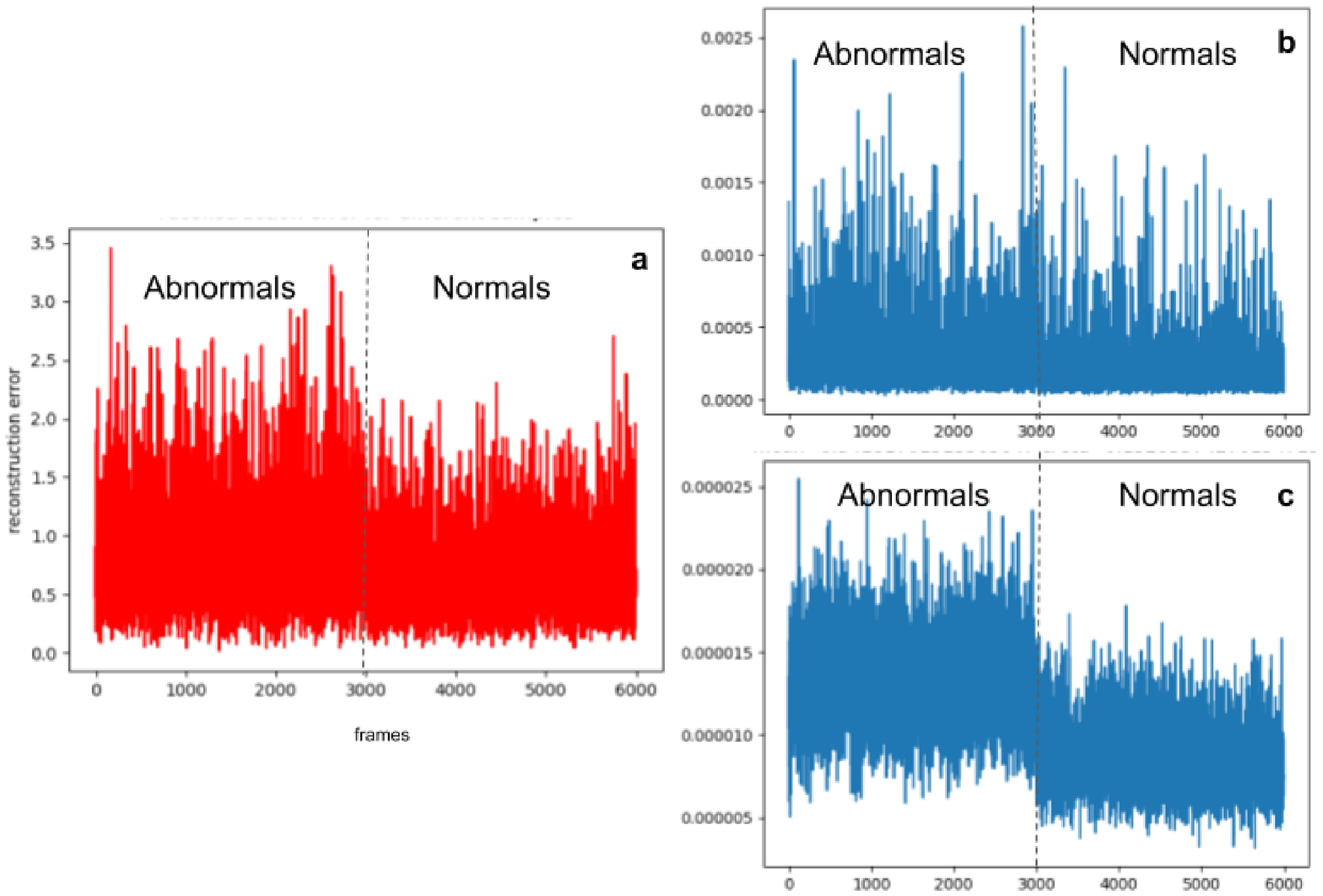}
    \caption{Reconstruction error for Normal and Abnormal samples. (a) Reconstruction error for images. (b) Reconstruction error for low-level features. (c) Reconstruction error for class-level features.}
    \label{fig:object-centric}
\end{figure*}

Results in Figure 26 show that the networks trained on images of the normal objects (Figure 26-a), or their low-level features (Figure 26-b) cannot effectively discriminate between normal and abnormal objects. Although frame reconstruction shows better results (in discriminating between normal and anomalies) compared to the low-level feature reconstruction method, it is not effective for anomaly detection. We believe that in frame reconstruction, the network implicitly considers some other features from the image, in addition to pixel intensity or color. However, the results of Figure 26-c show that training the network on class-level features (directly providing high-level features) helps the network effectively discriminate between normal and abnormal objects and improves the performance. As it can be concluded from this experiment, object-centric approaches do not effectively consider the class of the objects for discriminating between normal and abnormal objects.\\

Before summarizing the results, it is worth mentioning that these experiments do not target all proposed methods in the mentioned categories. However, they show that all these methods should consider the mentioned points (also summarized below) to reduce the false positives and false negatives. In summary, our experiments highlight these points:\\
\\
1)	Methods which focus on low-level features (in their loss or target functions) do not effectively discriminate between normal and abnormal frames which are defined based on the class of the object. These methods are also vulnerable to illumination changes.\\
2)	Methods which consider and analyze the frames entirely, instead of analyzing each object individually, will fail when the number of the foreground objects is considerably variable in different frames.\\
3)	Object-Centric methods are not affected by the number of foreground objects. However, they do not consider the environment information (BG information, location, etc.). Moreover, although they focus on objects (rather than redundant information such as BG), they are not aware of the class of the object if they reconstruct the cropped image in order to learn appearance patterns.\\
4)	It is probable that in single-path methods, which analyze appearance and motion simultaneously, the motion information may be dominated by the appearance. Two branch approaches, analyze the appearance and motion separately; hence their effect can be applied separately for the task.\\
5)	The effect of the camera distance (or object size) should be considered in the methods. Objects closer to the camera usually have more effect on the frame anomaly score.

\begin{table*}[!hbt]
 \caption{Advantages and challenges of different DL-based semi-supervised AD methods.}

{\footnotesize
 \begin{adjustwidth}{-0.5cm}{}
\begin{tabular}{|p{4.5cm} ||p{5.9 cm}|p{5.9cm}|  }

\hline
\textbf{AD strategy} & \textbf{Strong points} & \textbf{shortcomings}\\
\hline
\hline

\textbf{Reconstruction based} & Implicitly aware of the environment. & *Unaware of the object class.\newline*The model would be generalized to abnormals.\\
\hline

\textbf{Prediction based} & Implicitly aware of the environment. &*Unaware of the object class.\newline*The model would be generalized to abnormals. \\
\hline

\textbf{Object centric methods} & Focuses on the objects. &*The performance is dependant on the object detection step.\newline*Unaware of the environment. \\
\hline

\textbf{Segmentation based} & Is aware of the class of the object. &*Only detects the unexpected objects and is unaware of motion features.\\
\hline

\textbf{Memorization based} &Model is not generalizable to abnormals.  & *Produced anomaly scores would be equal for different frames if their latent spaces are close to each other.\\
\hline 

\end{tabular}
 \end{adjustwidth}
 }

\end{table*}

\begin{table*}[!hbt]
 \caption{Some of the state-of-the-art DL-based semi-supervised AD methods.}

{\footnotesize
 \begin{adjustwidth}{0.5cm}{}
\begin{tabular}{|p{4.2cm} ||p{2.2cm}| p{1.8cm}|p{5.9cm}|  }

\hline
\textbf{State of the art methods} & \textbf{Strategy for AD} & \textbf{DNN used in the method} & \textbf{Special points}\\
\hline
\hline

\textbf{Hasan et al. \cite{b37}} & Reconstruction & Conv-AE & *The proposed model (Conv-AE) does not consider temporal patterns effectively.
\newline*Not aware of the class of the objects. 
\newline*Has difficulty detecting anomalies in small regions.
 \\
\hline

\textbf{Chong et al. \cite{b15}} & Reconstruction & Conv-LSTM-AE & *Considers the evolution of frames. 
\newline*Not aware of the class of the objects. 
\\
\hline

\textbf{Akcay et al. \cite{b53}} & Reconstruction & Conditional GAN & *In addition to minimizing the distance between input and reconstructed images, distance between latent spaces is also minimized.\\
\hline

\textbf{Ravanbakhsh \cite{b78}} & Reconstruction & GAN & *Benefits from generating optical flow images from raw-pixel frames and vice versa for AD.\\
\hline

\textbf{Liu et al. \cite{b3}} & Prediction & GAN (Unet for generator)  & *Not aware of the class of the objects.\\
\hline 

\textbf{Ionescu et al. \cite{b68} } & Object centric & AE & *Focuses on the objects but does not consider the environment. \newline*The performance is dependant on the performance of the object detector.
 \\
\hline

\textbf{Gong et al. \cite{b76}} & Memorization & AE & *Benefits from a memory module and hence AE, here, is not generalizable to anomalies. \\
\hline

\textbf{Park et al. \cite{b77}} & Memorization & AE & *Considers the diversity of normal patterns explicitly. \newline*AE, here, is not generalizable to anomalies.\\
\hline

\end{tabular}
 \end{adjustwidth}
 }

\end{table*}

\section{Conclusion}

In this survey, recent DL-based semi-supervised video anomaly detection approaches and different DNNs, considering their use in anomaly detection, are reviewed. As DNNs are the main tool for different parts of the task (e.g., feature extraction, decision making), the survey began with a study of DNNs. Different DNNs are reviewed and analyzed from different points of view, such as spatiotemporal feature extraction, pattern learning, and compatibility with different data types. Moreover, their applicability for different parts of anomaly detection methods are stated, providing some points regarding their special attributes and challenges. Hence, researchers can choose the most suitable DNN for different parts of their anomaly detection method, based on their approaches. In the final section, different anomaly detection methods are critically reviewed. First, the methods are categorized based on spatiotemporal feature extraction process and then the survey analyzed them based on the strategy they commonly used for anomaly detection. Moreover, almost all of the recent approaches and state-of-the-art methods in the field is covered in this review, thereby providing a global but comprehensive look on the field for researchers, by describing essentials, positives points, shortcomings and challenges of each categorization and approach, which can be the subject of future work. Tables 4 and 5 summarize different reviewed anomaly detection strategies and Table 6 presents some of the state-of-the art research in this field.\\

Effective detection of video anomalies (similar to other applications requiring a good understanding of videos) requires joint consideration of different requirements, such as, extracting effective appearance features, capturing motion and extracting effective temporal features, separately capturing and analyzing different moving objects in the scene (for most applications), considering the context and the environment information, etc. However, each proposed method has addressed only one or a few of the mentioned requirements and almost all methods are unable to effectively and  jointly consider all of the aspects. This issue should be addressed in future work by properly combining existing methods, considering their capabilities and the capabilities of different DNNs.

\ifCLASSOPTIONcompsoc
\else
  \section*{Acknowledgment}
\fi


\ifCLASSOPTIONcaptionsoff
  \newpage
\fi

\end{document}